\documentclass[runningheads]{llncs}
\usepackage{graphicx}
\usepackage{comment}
\usepackage{amsmath,amssymb} 
\usepackage{color}
\usepackage{subfig}
\usepackage[american]{babel}
\usepackage{url}
\usepackage{xcolor}
\usepackage{multirow}
\usepackage{booktabs} 
\usepackage{pifont}
\usepackage{threeparttable}

\begin{document}
\pagestyle{headings}
\mainmatter
\def\ECCVSubNumber{1359}  

\title{Latency-Aware Differentiable Neural Architecture Search} 

\titlerunning{Latency-Aware Differentiable Neural Architecture Search} 
\authorrunning{Y. Xu \emph{et al.}} 
\author{%
	Yuhui Xu$^{1 {\ast}}$\quad Lingxi Xie$^{2}$\quad Xiaopeng Zhang$^{2}$\quad Xin Chen$^{3}$\quad Bowen Shi$^{1}$\\ Qi Tian$^{2}$\quad Hongkai Xiong$^{1}$}
\institute{$^{1}$Shanghai Jiao Tong University \quad $^{2}$Huawei Noah's Ark Lab \quad $^{3}$Tongji University}

\maketitle

\begin{abstract}\let\thefootnote\relax\footnote{$^{\ast}$ This work was done when Yuhui Xu and Xin Chen were interns at Huawei Noah's Ark Lab.}
Differentiable neural architecture search methods became popular in recent years, mainly due to their low search costs and flexibility in designing the search space. However, these methods suffer the difficulty in optimizing network, so that the searched network is often unfriendly to hardware. This paper deals with this problem by adding a differentiable latency loss term into optimization, so that the search process can tradeoff between accuracy and latency with a balancing coefficient. The core of latency prediction is to encode each network architecture and feed it into a multi-layer regressor, with the training data which can be easily collected from randomly sampling a number of architectures and evaluating them on the hardware. We evaluate our approach on NVIDIA Tesla-P100 GPUs. With 100K sampled architectures (requiring a few hours), the latency prediction module arrives at a relative error of lower than 10\%. Equipped with this module, the search method can reduce the latency by 20\% meanwhile preserving the accuracy. Our approach also enjoys the ability of being transplanted to a wide range of hardware platforms with very few efforts, or being used to optimizing other non-differentiable factors such as power consumption.
\end{abstract}

\section{Introduction}

Neural architecture search (NAS) is an important topic in an emerging research field named automated machine learning (AutoML). The idea is to design automatic algorithms to explore a complicated space which contains a very large number of network architectures and find out the best one(s) among them. Existing NAS algorithms are roughly categorized into two parts~\cite{Elsken2018NeuralAS,Wistuba2019ASO}, namely, heuristic search and differentiable search, differing from each other in whether the processes of sampling network from the space and training the sampled network are jointly optimized. Often, heuristic NAS methods (including using reinforcement learning~\cite{zoph2016neural,zoph2018learning,liu2018progressive} or genetic algorithms~\cite{real2017large,xie2017genetic,real2018regularized} for heuristic sampling) are computationally challenging caused by training sampled networks repeatedly, while differentiable NAS methods~\cite{liu2018darts,cai2018proxylessnas} are faster due to a larger fraction of shared training among sampled architectures.

Besides recognition accuracy, efficiency is also a pursuit of many real-world scenarios. This often requires the searched architecture to have a low \textbf{latency} at the inference time. For this respect, it is straightforward to undergo a multi-target training scheme in which accuracy and latency get optimized together. This is easy for heuristic search methods~\cite{tan2018mnasnet,Wu2018FBNetHE,Howard2019SearchingFM}, however, relatively difficult for the differentiable counterparts since latency is non-differentiable with respect to network parameters, except for the scenarios that the search space is very simple, \textit{e.g.}, the networks are chain-style so that the latency can be obtained via a lookup table~\cite{Wu2018FBNetHE}.

\begin{figure}[!t]
\centering
\includegraphics[width=0.45\textwidth]{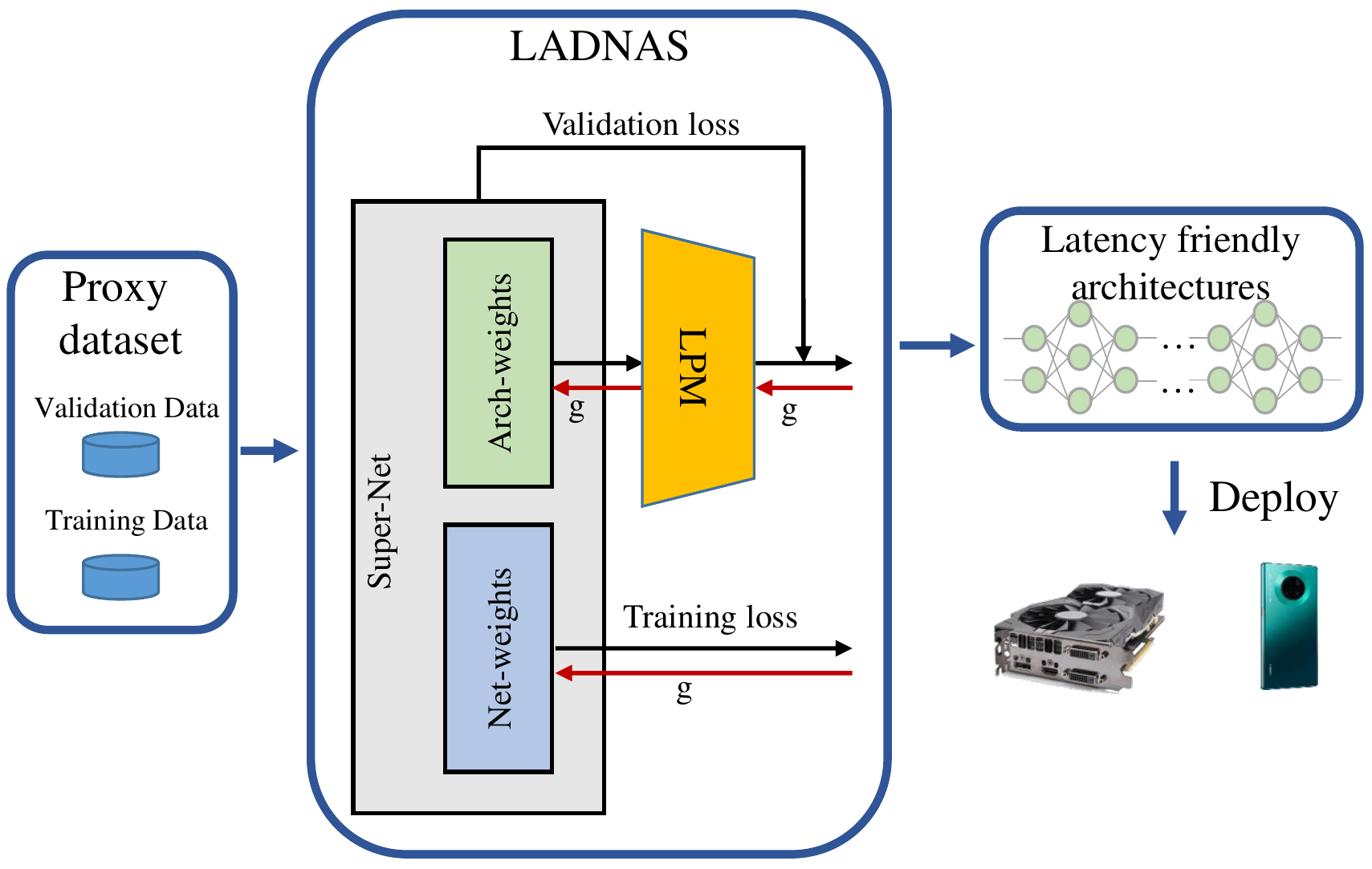}
\includegraphics[width=0.45\textwidth,height=4cm]{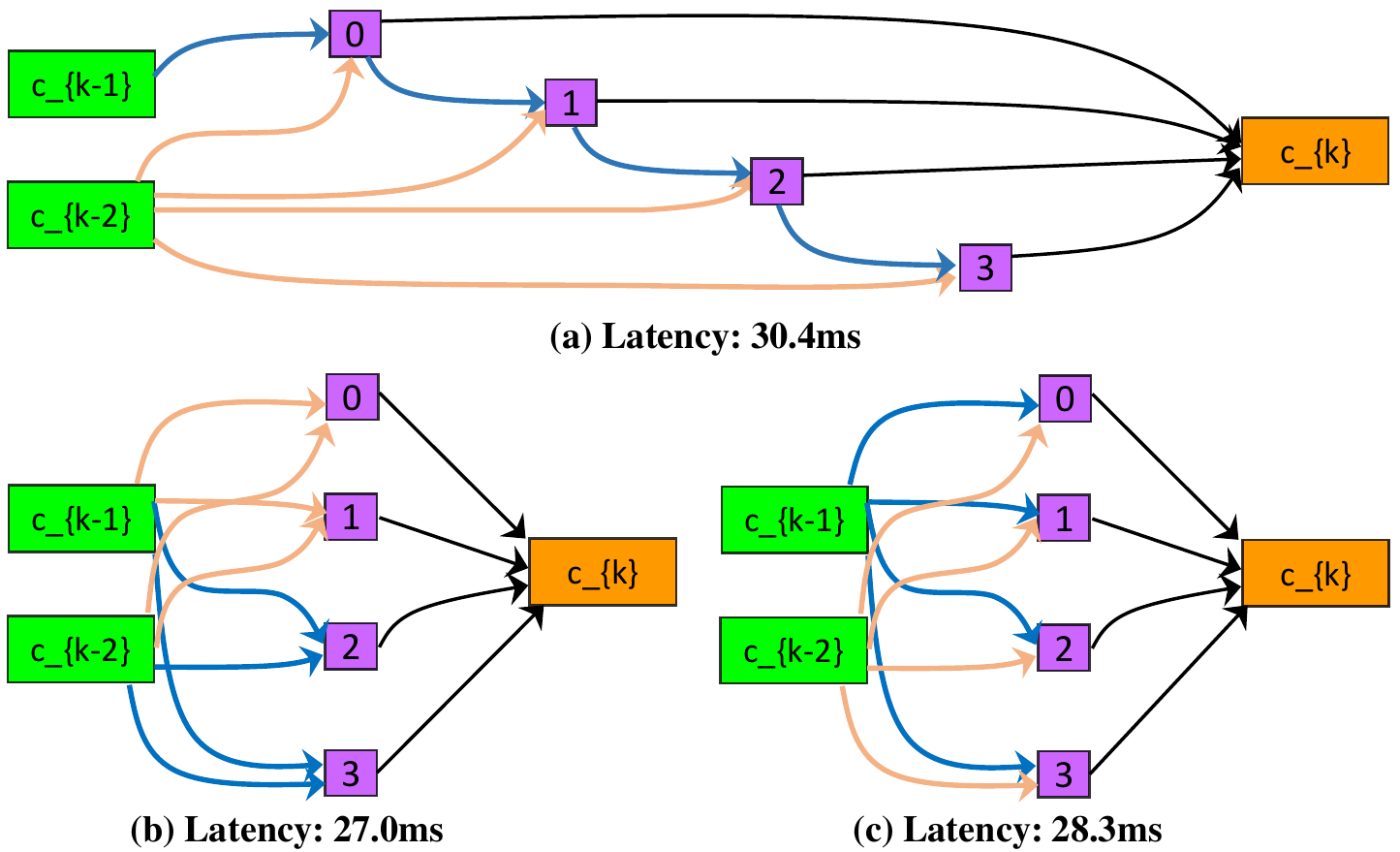}
\caption{\textbf{Left}: the goal of this paper is to introduce latency prediction to differentiable NAS methods towards a tradeoff between network performance and efficiency. \textbf{Right}: the latency of architectures in the DARTS space is difficult to predict, due to the potentially complex topology. Four \textsf{skip-connect} and four \textsf{sep-conv-3x3} operators can compose into different cells which have the same FLOPs but different latency values. The blue and yellow arrows in each cell indicate \textsf{skip-connect} and \textsf{sep-conv-3x3} operators, respectively}
\label{fig_introduction}
\end{figure}

This paper explores latency-aware differentiable architecture search \textbf{in a complicated space}, \textit{e.g.}, the DARTS~\cite{liu2018darts} space which contains a few nodes as well as topological connections between them, which exceeds the ability of table lookup. As shown in Figure~\ref{fig_introduction}, the relationship between latency and FLOPs of an architecture can be complex, and so it is unlikely to predict the latency with an empirically designed, arithmetic function with respect to the FLOPs.

Our idea is to train a differentiable \textbf{latency prediction module} (LPM) that is able to predict the latency of an architecture. LPM is a multi-layer neural network, with the input being an encoded form of an architecture, \textit{e.g.}, a fixed-length code of architectural parameters, and the output being the latency of the architecture. We train LPM by sampling a large number of architectures from the search space and measuring the latency of each of them. Note that though latency is closely related to the machine configuration, LPM is adaptive and can be trained for each specified hardware/software environment. In practice, we sampled $100\textrm{K}$ architectures from the DARTS space for training, which took around $9$ hours in a single NVIDIA Tesla-P100 GPU (the batch size is $32$), or $24$ hours in a Intel E5-1620 CPU (the batch size is $1$). The average relative error of latency prediction is smaller than $5\%$, which is (verified in experiments) accurate enough for our purpose, \textit{i.e.}, searching for latency-friendly architectures.

Equipped with LPM, we add the latency term to the loss function of DARTS. By setting different balancing coefficients, we can easily tradeoff between accuracy and speed, which is what we desire. We evaluate our approach on CIFAR10 and ImageNet, two standard image classification benchmarks. We arrive at similar classification accuracy with the baseline but our architecture is $15\%$--$20\%$ faster. In addition, our approach is easily transplanted to different hardware environments with acceptable costs. We train two LPM's on GPU and CPU, respectively, and they show different properties, \textit{i.e.}, the optimal architecture found on one device is often sub-optimal on another, demonstrating the need of hardware-specific architecture design.

The remainder of this paper is organized as follows. Section~\ref{relatedwork} briefly reviews the previous literature, and Section~\ref{approach} elaborates the algorithm for latency-aware architecture search. Experiments on both GPU and CPU are shown in Section~\ref{experiments}, and conclusions are drawn in Section~\ref{conclusions}.

\section{Related Works}
\label{relatedwork}

The past years have witnessed a rapid development of deep learning and manually-designed convolutional neural networks (CNNs) have pushed a wide range of computer vision tasks to new state-of-the-art performances~\cite{krizhevsky2012imagenet,simonyan2014very,he2016deep,huang2017densely}.
Lately, neural architecture search (NAS) has been attracting attentions due to its ability in automatically discovering network architectures with high performance. 

According to the methodology to explore the search space, existing NAS approaches can roughly be divided into two categories, namely, heuristic search and differentiable search. In some pioneer work in this area, architectures were sampled from the search space and trained from scratch to evaluate their capability, for which some heuristic algorithms, such as evolutionary algorithms and reinforcement learning, act as parameterized controllers of the sampling process. Among them, Liu~\textit{et al.}~\cite{liu2017hierarchical}, Xie~\textit{et al.}~\cite{xie2017genetic} and Real~\textit{et al.}~\cite{real2018regularized} adopted evolutionary algorithms as the controller, in which genetic operations were used to modify the architecture, and Real~\textit{et al.}~\cite{real2018regularized} showed that better evolutionary algorithms lead to stronger architectures. Another line of heuristics replaced evolutionary algorithms with reinforcement learning (RL)~\cite{zoph2016neural,baker2016designing,zoph2018learning,zhong2018practical,liu2018progressive}, in which a meta-controller is trained to generate the hyper-parameters of each candidate.

A crucial drawback of the above methods is the large search cost (hundreds or even thousands of GPU-days). In order to accomplish the search process with an acceptable cost, differentiable search methods were designed. In DARTS~\cite{liu2018darts}, Liu~\textit{et al.} introduced a set of architectural parameters to relax the search space so that the search process can be finished in a single training process, where the network parameters and the architectural parameters are jointly optimized and the final architecture is generated according to the architectural parameters. Following DARTS, ProxylessNAS~\cite{cai2018proxylessnas} adopted a similar differentiable framework and proposed to search architectures directly on the target dataset. To improve the stability of DARTS, P-DARTS~\cite{chen2019progressive} proposed to progressively enlarge the search depth to bridge the depth gap, and PC-DARTS~\cite{xu2019pc} enabled partial channel connection so that a large batch size can be used in the search process.

There also exist efforts in studying the hardware applicability of the discovered architecture in terms of FLOPs and/or latency. It is relatively easy for heuristic search methods to achieve this goal, because hardware constraints like FLOPs or latency can be conveniently measured for any sampled architecture~\cite{tan2018mnasnet,guo2019single}. Regarding differentiable NAS approaches, SNAS~\cite{xie2018snas} added FLOPs and memory access constraints by factorizing the architectural parameters and measuring the costs on each operation in the search space. ProxylessNAS~\cite{cai2018proxylessnas} and FBNet~\cite{Wu2018FBNetHE} adopted latency constraints since the search space is chain-styled and those constraints are accessible with a lookup table. To the best of our knowledge, no existing work has done the job in a complicated, differentiable search space, \textit{e.g.}, the search space of DARTS-based approaches.

\section{Approach}
\label{approach}

\subsection{DARTS and the Difficulty of Latency Prediction}
\label{approach:difficulty}

The goal of DARTS is to search for the robust cell architectures to construct the evaluation network. Specifically, a cell is represented by a directed acyclic graph (DAG) of $N$ nodes, $\{\mathbf{x}_0,\mathbf{x}_1,\ldots,\mathbf{x}_{N-1}\}$, where each node represents a set of feature maps. The first two nodes are the result feature maps of previous cells or operations and act as input nodes. Information flow between an intermediate node $j$ and its predecessor node $i$ is connected by an edge $E_{(i,j)}$, where a bunch of candidate operations $o(\cdot)$ in the operation space $\mathcal{O}$ are weighted by the normalized architectural parameters $\boldsymbol{\alpha}^{(i,j)}$, ${i}<{j}$, and formulated as:
\begin{equation}
f_{i,j}(\mathbf{x}_i) = \sum_{o\in\mathcal{O}_{i,j}}{\frac{\mathrm{exp}(\alpha_o^{(i,j)})}{\sum_{o'\in\mathcal{O}}\mathrm{exp}(\alpha_{o'}^{(i,j)})}o(\mathbf{x}_i)}.
\end{equation}
An intermediate node is the summation of the outputs of its preceding edges, which is represented as $\mathbf{x}_j=\sum_{i<j}{f_{i,j}(\mathbf{x}_i)}$, and the output node is the concatenation of all intermediate nodes in the channel dimension, which is denoted by $\mathbf{x}^\mathrm{output}=\mathrm{concat}(\mathbf{x}_2,\mathbf{x}_3,\ldots,\mathbf{x}_{N-1})$. In this manner, DARTS defines an over-parameterized network $\mathbf{h}\!\left(\mathbf{x};\boldsymbol{\omega},\boldsymbol{\alpha}\right)$ where $\boldsymbol{\omega}$ and $\boldsymbol{\alpha}$ denote the network and architectural parameters. With a bi-level optimization process, $\boldsymbol{\omega}$ and $\boldsymbol{\alpha}$ are trained in a proxy dataset and $\boldsymbol{\alpha}$ is used to determine the final architecture.

Despite the satisfying performance of the searched architecture, we are not sure if the architecture is also optimized in terms of efficiency, \textit{e.g.}, latency. In particular, DARTS involves many inter-layer connections (\textit{e.g.}, each cell receives input from two previous cells) which may bring memory access issues and slow down the architecture. More importantly, such a complex architecture brings uncertainty in latency estimation, because the cost of memory access is often difficult to measure, unlike that of a specific operator. Hence, summing up the latency of all layers (stored in a lookup table~\cite{Wu2018FBNetHE}) is no longer accurate.

\begin{figure}[!t]
\centering
\includegraphics[width=0.45\textwidth]{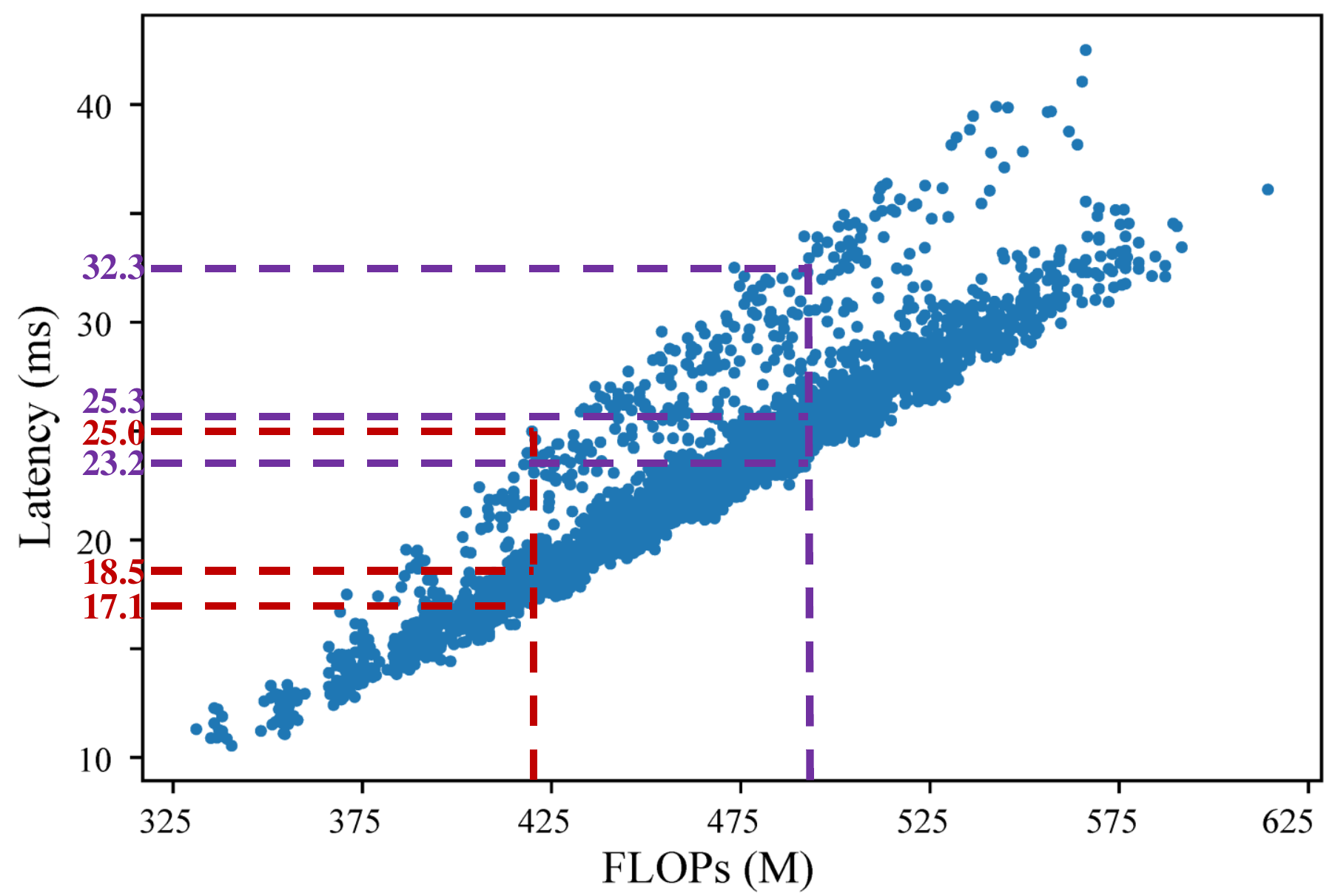}
\includegraphics[width=0.45\textwidth,height=4cm]{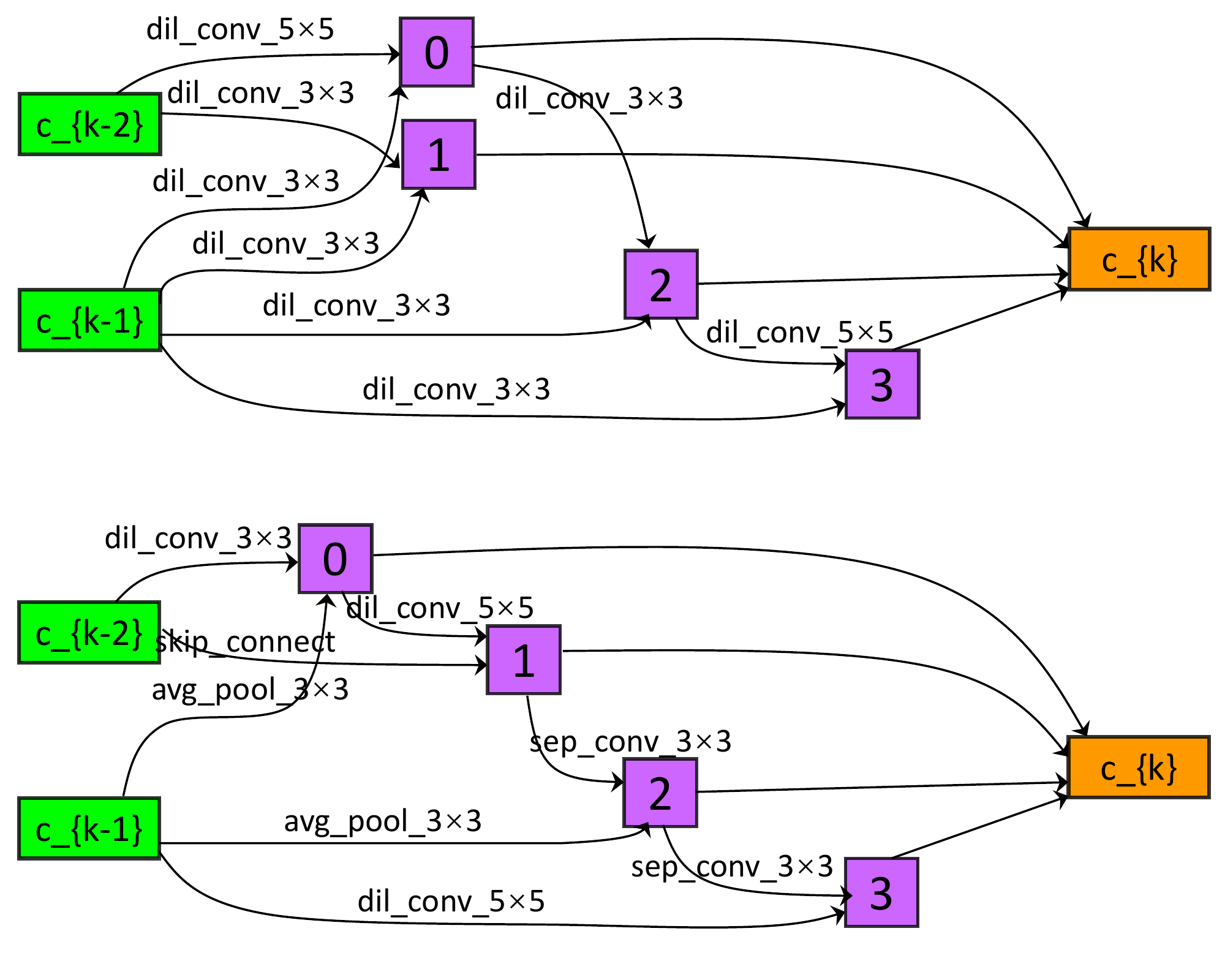}
\caption{We sample $10\mathrm{K}$ architectures from the DARTS space and plot the FLOPs as well as latency of each of them on ImageNet data ($224\times224$). \textbf{Left}: under a specified FLOPs, the smallest latency can be $32\%$ smaller than the largest one, or $8\%$ smaller than the median. \textbf{Right}: the slowest (top, $32.3\mathrm{ms}$) and fastest (bottom, $23.2\mathrm{ms}$) architectures (in normal cells) under $490\mathrm{M}$ FLOPs (the purple dashed line), in which the main difference is caused by the varying latency/FLOPs ratio, \textit{e.g.}, the \textsf{dil-conv-3x3} operator has nearly half FLOPs yet requires around $70\%$ latency compared to \textsf{sep-conv-3x3}}
\label{fig:difference}
\end{figure}

We verify this statement by observing the relationship between latency and FLOPs, which is closely related to the sum of latency of individual layers. As shown in Figure~\ref{fig:difference}, though the quantities of latency and FLOPs are positive related, the architectures of the same FLOPs can still have very different latency, with the fastest one being at least $30\%$ faster than the slowest one. That being said, FLOPs-aware search methods~\cite{xie2018snas} are not guaranteed to produce efficient results -- there is room for latency-aware search algorithms.

\subsection{Latency-Aware Differentiable Architecture Search}
\label{approach:pipeline}

We present a search framework which we call latency-aware differentiable neural architecture search (LA-DNAS). In particular, this paper follows the search space and optimization methods of DARTS, so we name our models \textbf{LA-DARTS}.

The key of LA-DARTS is to design a differentiable loss function that can predict the latency of the architecture parameter, $\boldsymbol{\alpha}$, so that it can be integrated into the over-parameterized network optimization process. We denote this function as $\mathrm{LAT}\!\left(\boldsymbol{\alpha}\right)$, which is the expectation of latency when an architecture is sampled according to the weights of $\boldsymbol{\alpha}$:
\begin{equation}
{\mathrm{LAT}\!\left(\boldsymbol{\alpha}\right)}={\mathbb{E}\!\left[\mathrm{LAT}\!\left(\boldsymbol{\gamma}\right),\boldsymbol{\gamma}\sim\mathcal{S}\!\left(\tilde{\boldsymbol{\alpha}}\right)\right]},
\end{equation}
where $\boldsymbol{\gamma}$ denotes a discretized sub-architecture that DARTS allows to appear, and $\mathcal{S}\!\left(\tilde{\boldsymbol{\alpha}}\right)$ denotes that sampling process is parameterized by $\tilde{\boldsymbol{\alpha}}$ ($\tilde{\boldsymbol{\alpha}}$ is the probabilistic values obtained by passing each edge of $\boldsymbol{\alpha}$ through softmax). In practice, we uniformly sample $8$ out of $14$ edges from $\tilde{\boldsymbol{\alpha}}$, and then randomly choose the operation on each edge according to the current weights of the operations (excluding \textsf{none} which does not appear in the final architecture). We use a batch size of ${M}={20}$, sample $M$ sub-architectures, $\left\{\boldsymbol{\gamma}_m\right\}_{m=1}^M$, and thus have ${\mathrm{LAT}\!\left(\boldsymbol{\alpha}\right)}\approx{\frac{1}{M}\sum_{m=1}^M\mathrm{LPM}\!\left(\boldsymbol{\gamma}_m\right)}$, where $\mathrm{LPM}\!\left(\cdot\right)$ denotes a latency prediction function which will be detailed in the next subsection. The final loss function of the search process is written as:
\begin{equation}
{\mathcal{L}_\mathrm{total}\!\left(\boldsymbol{\alpha}\right)}={\mathcal{L}_\mathrm{val}\!\left(\boldsymbol{\alpha}\right)+\lambda\cdot\mathrm{LAT}\!\left(\boldsymbol{\alpha}\right)}.
\end{equation}
Here, the balancing coefficient, $\lambda$, controls the tradeoff between accuracy and performance: a smaller $\lambda$ prefers accuracy to latency and vice versa. Note that $\lambda$ has a unit of $\mathrm{sec}^{-1}$. We will show in experiments that choosing a proper $\lambda$ is not difficult, and adjusting $\lambda$ can lead to different properties of architectures. Upon the differentiability of $\mathrm{LAT}\!\left(\boldsymbol{\alpha}\right)$, this loss function is easily optimized following the bi-level optimization of DARTS-based approaches.

To compute the gradient of $\mathrm{LAT}\!\left(\boldsymbol{\alpha}\right)$ with respect to $\boldsymbol{\alpha}$, we have:
\begin{equation}
{\frac{\partial\mathrm{LAT}\!\left(\boldsymbol{\alpha}\right)}{\partial\boldsymbol{\alpha}}}\approx{\frac{1}{M}\sum_{m=1}^M\frac{\partial\mathrm{LAT}\!\left(\boldsymbol{\gamma}_m\right)}{\partial\boldsymbol{\gamma}_m}\cdot\frac{\partial\boldsymbol{\gamma}_m}{\partial\tilde{\boldsymbol{\alpha}}}\cdot\frac{\partial\tilde{\boldsymbol{\alpha}}}{\partial\boldsymbol{\alpha}}}\approx{\frac{1}{M}\sum_{m=1}^M\frac{\partial\mathrm{LAT}\!\left(\boldsymbol{\gamma}_m\right)}{\partial\boldsymbol{\gamma}_m}\cdot\frac{\partial\tilde{\boldsymbol{\alpha}}}{\partial\boldsymbol{\alpha}}}.
\end{equation}
Here, as $\boldsymbol{\gamma}_m$ is the binarization of $\tilde{\boldsymbol{\alpha}}$, we use the straight-through gradient estimator~\cite{bengio2013estimating}, the gradient goes “straight-through” $\boldsymbol{\gamma}_m$, so that  ${\partial\boldsymbol{\gamma}_m/\partial\tilde{\boldsymbol{\alpha}}}\approx{\mathbf{I}}$.

\begin{figure}[!t]
\centering
\includegraphics[width=0.90\textwidth]{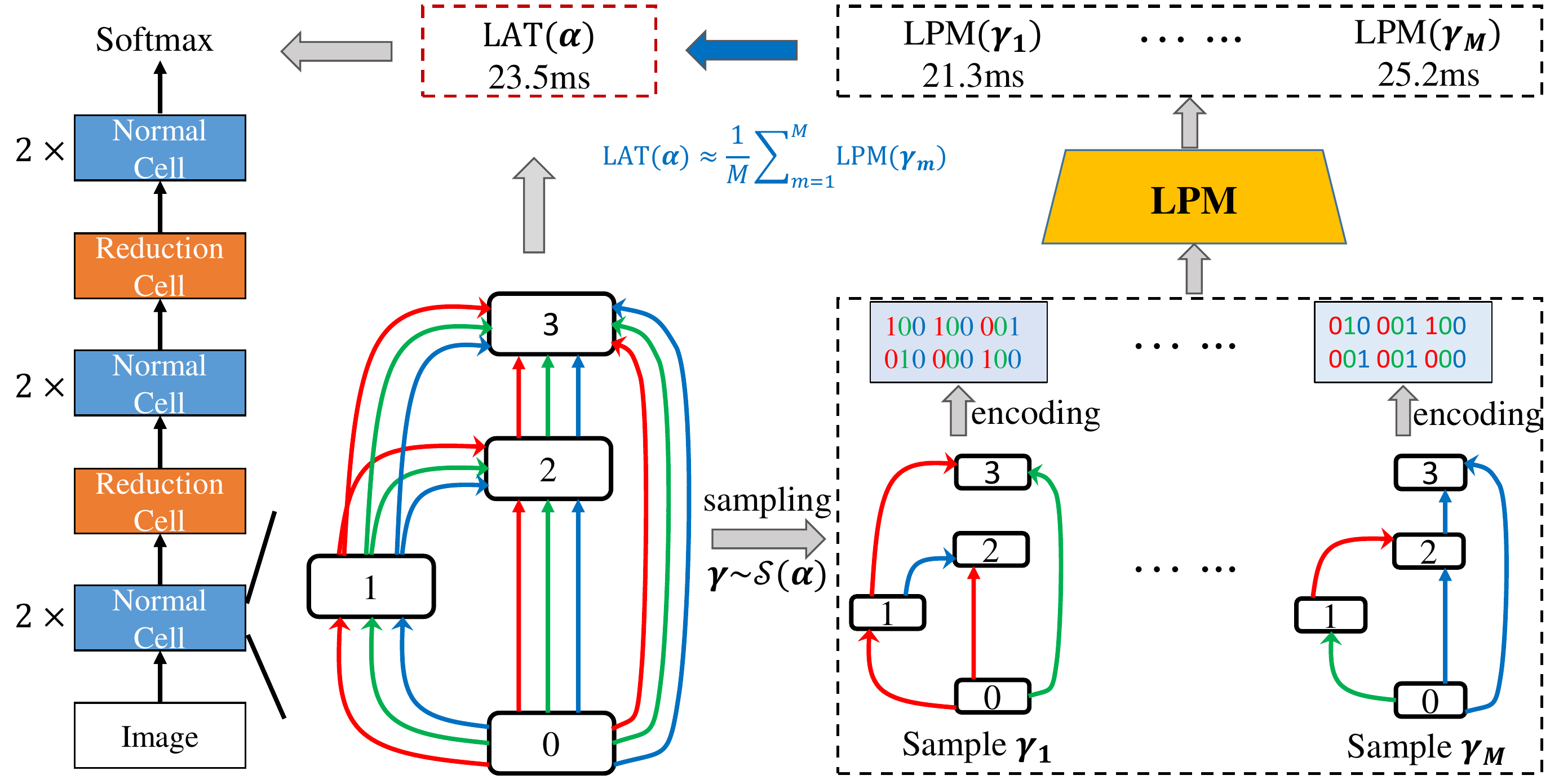}\\
\caption{Illustration of the proposed latency-aware differentiable architecture search (best viewed in color). The latency of the current over-parameterized network is estimated by sampling sub-networks from it, feeding them into the pre-trained latency prediction module (LPM), and averaging the results. The binary code indicates the encoded architectures, in which we use a simplified super-network with ${\left|\mathcal{O}\right|}={3}$ for better visualization (in DARTS, ${\left|\mathcal{O}\right|}={8}$)}
\label{fig_network}
\end{figure}

The overall pipeline of our approach is illustrated in Figure~\ref{fig_network}.

\subsection{Training a Latency Prediction Module}
\label{approach:latency}

It remains to design a \textbf{latency prediction module} (LPM) which outputs a value of $\mathrm{LPM}\!\left(\boldsymbol{\gamma}_m\right)$ for each sampled sub-architecture, $\boldsymbol{\gamma}_m$. We present a learning-based solution for the following reasons. First, we believe that latency is learnable. In other words, there exist network architecture patterns that correspond to latency, so that a deep network can learn to predict with sufficient training data). Second, latency prediction does not need to be very accurate, small errors are acceptable (in the experimental section, we verify that the error of our prediction is sufficiently small, and more importantly, small inaccuracy barely harms search performance). Third, we can easily transplant the learning-based approach to other device without much expertise which eases the deployment of NAS on a wide range of hardware. We will show an example in Section~\ref{experiments:CPU}.

$\mathrm{LPM}\!\left(\boldsymbol{\gamma}_m\right)$ is a multi-layer regression network, with the input being an encoded \textbf{sub-architecture} and the output being the predicted latency. Throughout this paper, we only investigate the normal cell and ignore the reduction cell, because the final reduction cell is often composed of weight-free operators and contributes little to the network latency. On the other hand, encoding the reduction cell introduces noise to the latency prediction model.

To encode the sub-architecture, we first recall that each cell of DARTS contains four intermediate nodes with $14$ edges and $8$ operations on each edge, while the sub-architecture preserves two edges for each node and only one operation on each selected edge. We use $14\times8$ bits to represent each cell: a bit is $1$ if it corresponds to the chosen operation on a preserved edge, otherwise it is $0$. In other words, only $8$ out of $14\times8$ bits are $1$. The $112$D vector is propagated through four fully-connected layers with $112$, $256$, $64$ and $1$ neurons, respectively, and the final one is the output (latency). We use \textsf{sigmoid} as the activation function for each layer, excluding the last one.

\textbf{Data collection.} We first collect a dataset of (architecture, latency) pairs. On an NVIDIA Tesla-P100 GPU (used in all experiments of our work), we randomly sample $100\mathrm{K}$ architectures from the DARTS space, and evaluate the latency of each architecture with randomized network weights. For a better transferability of the searched architectures, the latency is measured under the ImageNet setting with an input image size of $224\times224$ and is an average of $20$ measurements. The entire process takes around $9$ hours. We also evaluate the latency for the same set of architectures on an Intel E5-1620 CPU, which takes $24$ hours. Though $100\mathrm{K}$ is a small number compared to the entire search space (there are $1.0\times10^9$ distinct normal cells), it is enough for the learning task. Then we partition the latency data into two parts: $80\mathrm{K}$ pairs are used for training and the remaining $20\mathrm{K}$ for validation.

\begin{table}[!t]
\begin{center}
\caption{Absolute and relative errors of the LPM over the $20\mathrm{K}$ testing architectures when using different numbers of training architectures. On GPU and CPU, we sample the same set of $100\mathrm{K}$ architectures and use the same data split}
\label{table.1}
\setlength{\tabcolsep}{0.12cm}
\begin{tabular}{|l||c|c|c|c|c||c|c|c|c|c|}
\hline
{} & \multicolumn{5}{c||}{NVIDIA Tesla-P100 GPU} & \multicolumn{5}{c|}{Intel E5-1620 CPU} \\
\hline
Training Data & $10\mathrm{K}$  & $20\mathrm{K}$  & $40\mathrm{K}$  & $60\mathrm{K}$  & $80\mathrm{K}$ & $10\mathrm{K}$  & $20\mathrm{K}$  & $40\mathrm{K}$  & $60\mathrm{K}$  & $80\mathrm{K}$ \\
\hline\hline
Absolute Error (ms) & 1.79 & 1.41 & 1.09 & 0.84 & 0.82 & 20.24 & 16.64 & 10.42 & 8.50 & 8.27 \\
\hline
Relative Error (\%) & 8.09 & 6.23 & 4.79 & 3.57 & 3.45 & 12.13 & 10.10 & 7.69 & 5.79& 5.32 \\
\hline
\end{tabular}
\end{center}
\end{table}

\textbf{Training and inference.} On the $80\mathrm{K}$ training set, the network is trained from scratch for $1\rm{,}000$ epochs using a batch size of $200$. We use a momentum SGD with a fixed learning rate of $0.01$, a momentum of $0.9$, a weight decay of $1\times10^{-5}$, and a mean square error (MSE) loss function. We evaluate LPM using both absolute and relative errors between the prediction and the ground-truth on the testing set. As shown in Table~\ref{table.1}, with an increasing amount of training data, the testing error goes down accordingly. On the other hand, the improvement of accuracy becomes marginal when the amount of training data is larger than $60\mathrm{K}$. With $80\mathrm{K}$ training data, the latency prediction results are satisfying, with an absolute error smaller than $1\mathrm{ms}$ on GPU or smaller than $10\mathrm{ms}$ on CPU, and a relative error smaller than $4\%$ on GPU or smaller than $4\%$ on CPU.

To further show the consistency between the ground-truth and predicted latency values, we also sample $2\mathrm{K}$ architectures from the testing set and compute the Kendall-$\tau$ coefficient. The $\tau$-value is $0.83$ for GPU and $0.75$ for CPU, indicating that $92\%$ and $87\%$ architecture pairs have the same relative ranking in the ground-truth and predicted lists. As we shall see in experiments, such accuracy is sufficient in finding efficient yet powerful architectures.

\subsection{Discussions and Relationship to Prior Works}
\label{approach:discussions}

To the best of our knowledge, this is the first work that introduces a latency-aware method to a complicated search space. The main difficulty lies in designing a differentiable loss function for latency prediction, while this issue does not exist for heuristic search methods. There are a lot of efforts in applying latency constraints to heuristic search~\cite{tan2018mnasnet,Wu2018FBNetHE,Howard2019SearchingFM}. 

On the other hand, in differentiable architecture search, FBNet~\cite{Wu2018FBNetHE} which integrated latency into the loss function by constructing a look-up table. Although this method works well in the chain-style search space, it can fail in the search space of DARTS due to much higher complexity. In comparison, our approach has a stronger ability and is feasible for a wider range of search spaces. Also, there were efforts~\cite{xie2018snas} in introducing naturally differentiable quantities, \textit{e.g.}, FLOPs (a linear function of $\boldsymbol{\alpha}$, to the loss function of differentiable frameworks. Our approach, in comparison, is more generalized.

\section{Experiments}
\label{experiments}

We evaluate our approach on two standard image classification benchmarks, \textit{i.e.}, CIFAR10 and ImageNet, to study several important properties of it. We first use the latency prediction on an NVIDIA Tesla-V100 GPU, and then generalize it to that on an Intel E5-1620 CPU.

\subsection{Experiments on CIFAR10}
\label{experiments:CIFAR10}

Firstly, we evaluate our LADNAS on CIFAR10~\cite{krizhevsky2009learning}. The CIFAR10 dataset
consists of 60k colored natural images with 32$\times$32 resolution of 10 categories, which is split into 50K training and 10K testing images. We use DARTS~\cite{liu2018darts} and PC-DARTS~\cite{xie2018snas} as our two baseline methods. Following DARTS and PC-DARTS, we use an individual stage for architecture search and conduct another standalone training process from scratch to evaluate the optimal architecture obtained in the search phase. In the search stage, the goal is to determine the best sets of architectural parameters, namely $\left\{\alpha_{i,j}^o\right\}$ in DARTS and $\left\{\alpha_{i,j}^o\right\}$, $\left\{\beta_{i,j}\right\}$ in PC-DARTS for each edge $E_{(i,j)}$. To this end, the training
set is partitioned into two parts, with the first part used for optimizing network parameters, \textit{e.g.},
convolutional weights, and the second part used for optimizing architectural parameters. For fair comparison, the operation space $\mathcal{O}$ remains the same as the convention, which contains $8$ choices, \textit{i.e.}, \textsf{sep-conv-3x3}, \textsf{sep-conv-5x5}, \textsf{dil-conv-3x3}, \textsf{dil-conv-5x5}, \textsf{max-pool-3x3}, \textsf{avg-pool-3x3}, \textsf{skip-connect} (\textsf{identity}), and \textsf{zero} (\textsf{none}).

Following DARTS and PC-DARTS, in the search period, the over-parameterized network is constructed by stacking $8$ cells ($6$ normal cells and $2$ reduction cells, each type of cells share the same architecture), and each cell consists of ${N}={6}$ nodes. We train the network for $50$ epochs, with the initial number of channels being $16$. In the search phase, the network weights are optimized by momentum SGD, with a batch size of $64$ for DARTS and $256$ for PC-DARTS, an initial learning rate of $0.025$ for DARTS and $0.1$ for PC-DARTS (annealed down to zero following the cosine schedule without restart), a momentum of 0.9, and a weight decay of $3\times10^{-4}$. We use an Adam optimizer~\cite{kingma2014adam} for architectural parameters, with a fixed learning rate of $3\times10^{-4}$ for DARTS and $6\times10^{-4}$ for PC-DARTS, a momentum of $(0.5,0.999)$ and a weight decay of $10^{-3}$. For PC-DARTS, we freeze architectural parameters and only allow network parameters to be tuned in the first $15$ epochs. For P-DARTS~\cite{chen2019progressive}, we add the the proposed modeule in the last search stage.

\begin{table}[!t]
\centering
\caption{Comparison with state-of-the-art network architectures on CIFAR10. Latency is measured on an NVIDIA Tesla-P100 GPU with a batch size of $32$ and an input size of $32\times32$. In latency-aware approaches, training the LPM requires additional $0.4$ GPU-days}
\label{tab.cifar}
\small
\begin{threeparttable}[b]
\resizebox{\textwidth}{!}{
\begin{tabular}{lcccccc}
\hline
\textbf{\multirow{2}{*}{Architecture}} & \textbf{Test Err.} & \textbf{Params} &\textbf{Latency}& \textbf{Search Cost} & \textbf{\multirow{2}{*}{Search Method}} \\
&                            \textbf{ (\%)} & \textbf{(M)}& \textbf{(ms)} & \textbf{(GPU-days)} &\\
\hline
DenseNet-BC~\cite{huang2017densely}                      & 3.46         & 25.6 &- & -    & manual \\
\hline
NASNet-A~\cite{zoph2018learning} + cutout                & 2.65         & 3.3  &- & 1800 & RL      \\
AmoebaNet-A~\cite{real2018regularized} + cutout          & 3.34$\pm$0.06 & 3.2 & -& 3150 & evolution \\
AmoebaNet-B~\cite{real2018regularized} + cutout          & 2.55$\pm$0.05 & 2.8  &- & 3150 & evolution \\
Hireachical Evolution~\cite{liu2017hierarchical}         & 3.75$\pm$0.12 & 15.7 &- & 300  & evolution \\
PNAS~\cite{liu2018progressive}                           & 3.41$\pm$0.09 & 3.2  &- & 225  & SMBO \\
ENAS~\cite{pham2018efficient} + cutout                   & 2.89          & 4.6  &- & 0.5  &RL \\
NAONet-WS~\cite{luo2018neural}                           & 3.53          & 3.1  &- & 0.4  & NAO\\
\hline
SNAS (mild)~\cite{xie2018snas} + cutout                  & 2.98            & 2.9  &30.2 & 1.5  & gradient-based \\
ProxylessNAS~\cite{cai2018proxylessnas} + cutout         & 2.08            & -    &- & 4.0 & gradient-based \\

BayesNAS~\cite{zhou2019bayesnas} + cutout& 2.81$\pm$0.04   & 3.4  &- & 0.2 & gradient-based \\
GDAS~\cite{dong2019searching} + cutout                   & 2.93            & 3.4  & 30.6& 0.3 & gradient-based \\
\hline
DARTS (2nd order)~\cite{liu2018darts} + cutout           & 2.76$\pm$0.09   &  3.3 & 40.9& 0.3 & gradient-based \\
LA-DARTS (2nd order) + cutout                            & 2.72$\pm$0.05   &  2.7 & 28.4& 0.3$+$0.4 & gradient-based \\
\hline
P-DARTS~\cite{chen2019progressive} + cutout              & 2.50            & 3.4  &40.9 & 0.3 & gradient-based \\
LA-P-DARTS + cutout & 2.52$\pm$0.08  & 3.3  &35.8 & 0.3$+$0.4 & gradient-based\\
\hline
PC-DARTS~\cite{xu2019pc} + cutout                        & 2.57$\pm$0.07   & 3.6  & 40.7& 0.1 & gradient-based \\
LA-PC-DARTS + cutout                                     & 2.61$\pm$0.10    &  2.6& 27.7 & 0.1$+$0.4 & gradient-based \\
\hline
\end{tabular}
}
\end{threeparttable}
\end{table}

\vspace{0.2cm}
\noindent$\bullet$\quad\textbf{Evaluation on CIFAR10}

The evaluation scenario simply follows that of DARTS and PC-DARTS. The evaluation network is stacked by $20$ cells ($18$ normal cells and $2$ reduction cells). The initial number of channels is $36$. The entire $50\mathrm{K}$ training set is used, and the network is trained from scratch for $600$ epochs using a batch size of $128$. We use the SGD optimizer with an initial learning rate of $0.025$ (annealed down to zero following a cosine schedule without restart), a momentum of $0.9$, a weight decay of $3\times10^{-4}$ and a norm gradient clipping at $5$. Drop-path with a rate of $0.2$ as well as cutout~\cite{devries2017improved} is also applied for regularization. The balancing coefficient $\lambda$ is set as $0.2$. The GPU latency on CIFAR10 is measured on one Tesla-P100 GPU with a batch size of $32$ (input image size $32\times32$) and is the average of $200$ measurements.

We conduct latency-aware architecture search on DARTS, P-DARTS, and PC-DARTS. As demonstrated in Table~\ref{tab.cifar}, LA-DARTS (2nd order) achieves a $2.72\%$ test error with only $2.7\mathrm{M}$ parameters and a latency of $28.4\mathrm{ms}$ on CIFAR10. To achieve a similar classification performance, the original DARTS (2nd order) need $3.3\mathrm{M}$ parameters with $40.9\mathrm{ms}$ latency. SNAS~\cite{xie2018snas} can obtain relative good latency by mild FLOPs constraint, however, this strict constraint leads to a much worse performance. Compared to P-DARTS and PC-DARTS, the latency-aware variants of them report nearly the same performance but with $10\%$ and $30\%$ relative drop in latency, respectively.

\begin{figure}[t]
\centering
\begin{minipage}{0.49\textwidth}
\subfloat[$\lambda=0.00$, Lat.: $40.7\mathrm{ms}$, Err.: $2.57\%$]{\includegraphics[width=1.0\linewidth]{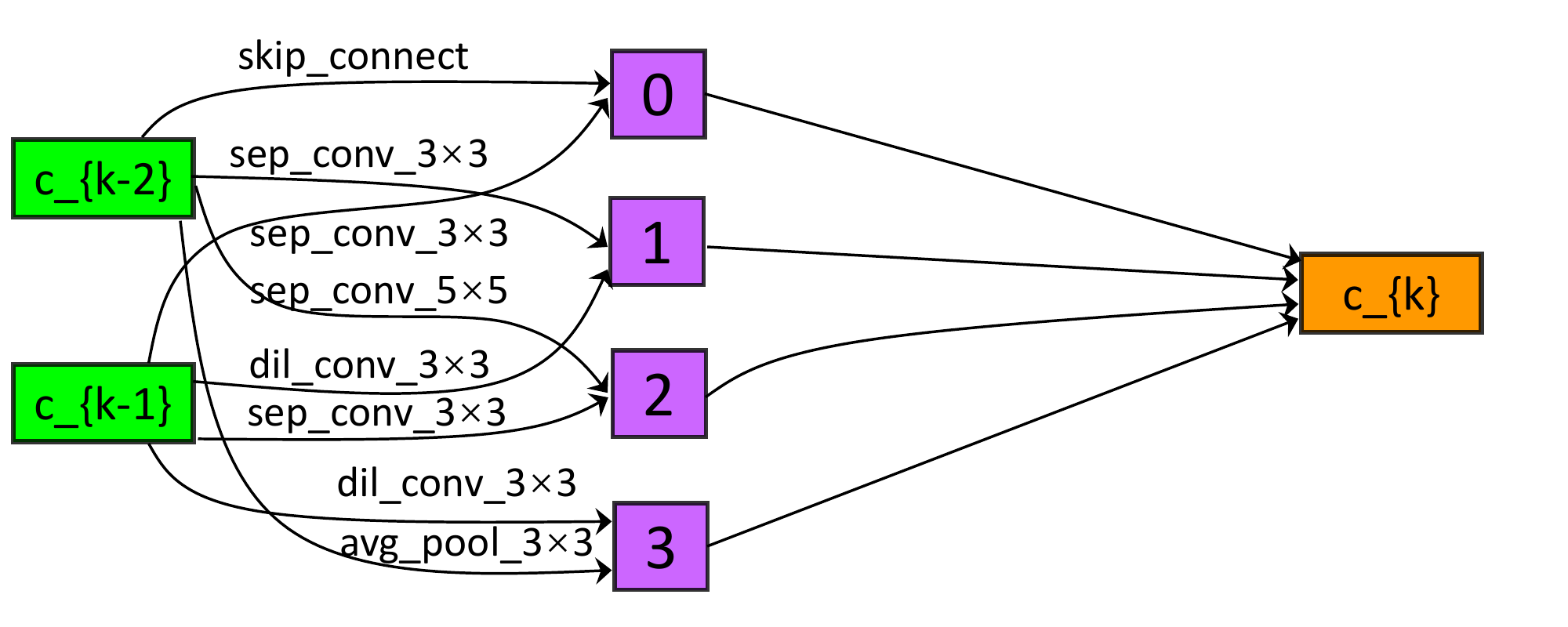}}\label{ncells_s1}\\
\subfloat[$\lambda=0.10$, Lat.: $35.5\mathrm{ms}$, Err.: $2.64\%$]{\includegraphics[width=1.0\linewidth]{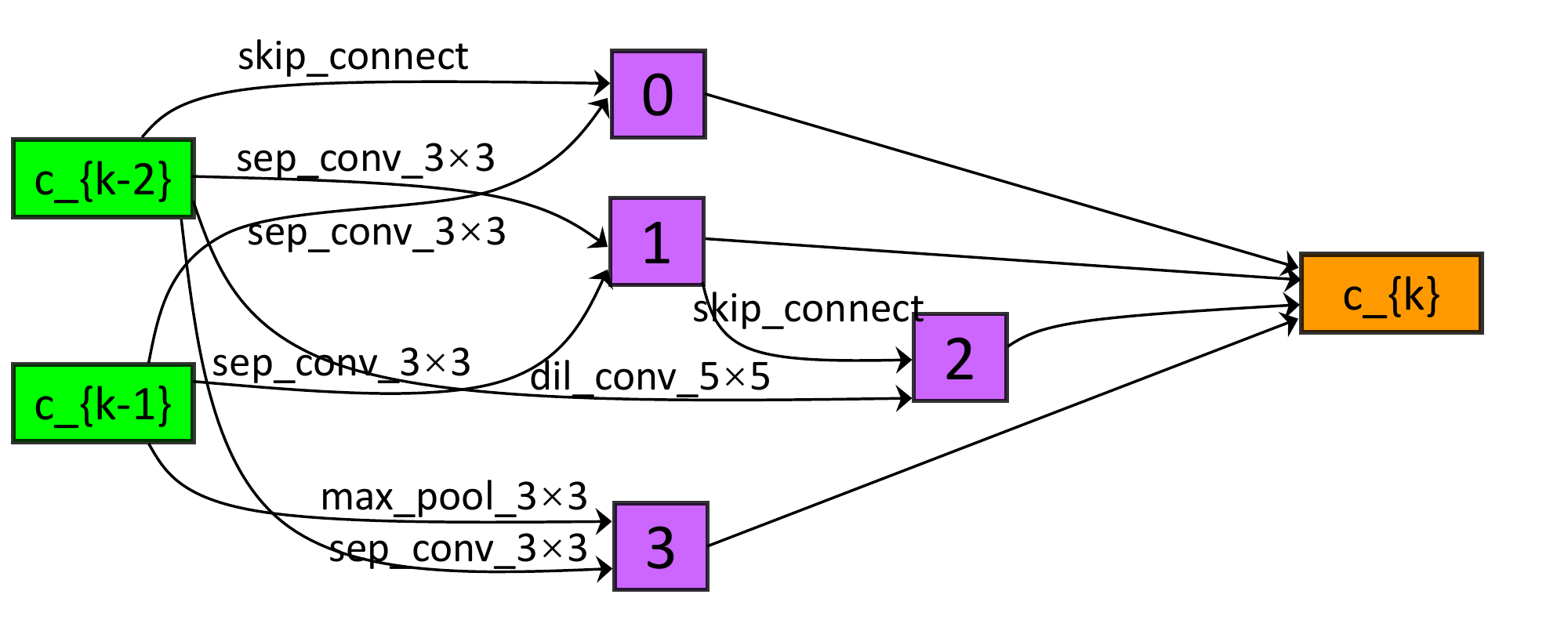}}\label{ncells_s2}
\end{minipage}
\begin{minipage}{0.49\textwidth}
\subfloat[$\lambda=0.15$, Lat.: $31.2\mathrm{ms}$, Err.: $2.69\%$]{\includegraphics[width=1.0\linewidth]{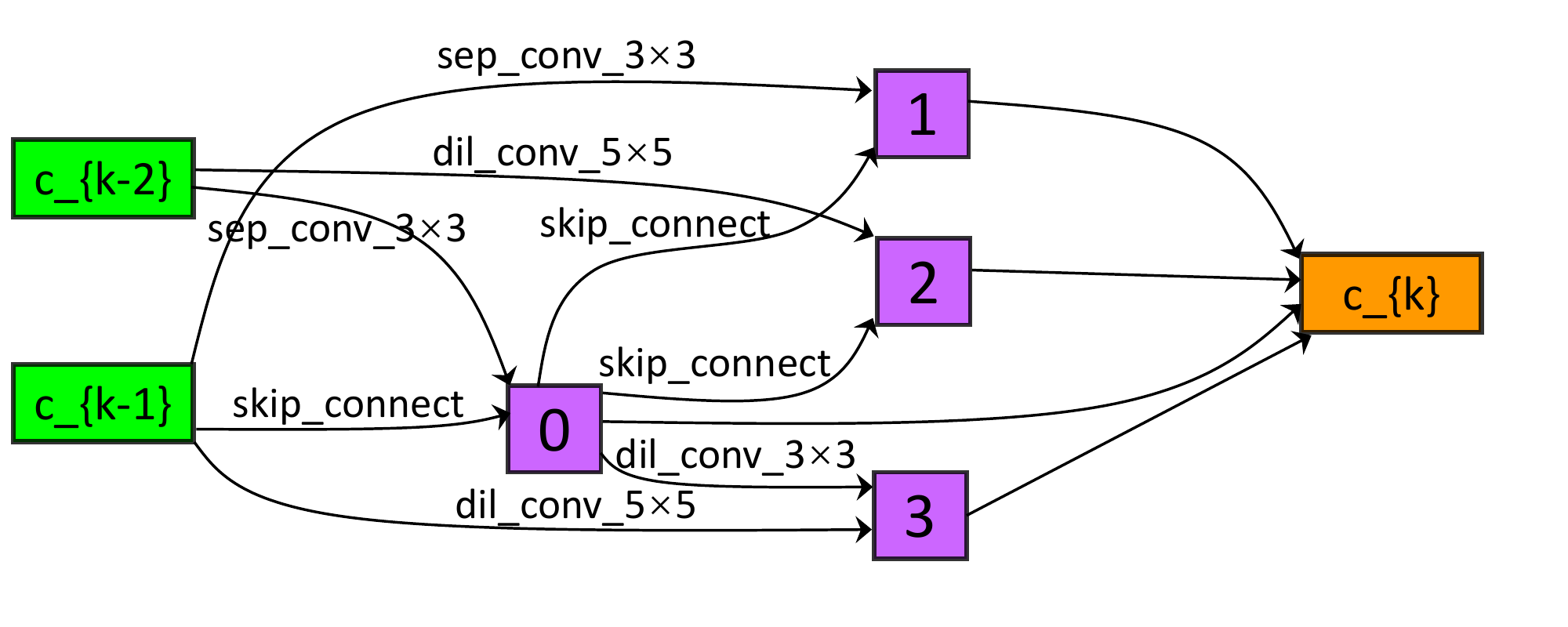}}\label{ncells_s3}\\
\begin{center}
\subfloat[$\lambda=0.20$, Lat.: $27.7\mathrm{ms}$, Err.: $2.61\%$]{\includegraphics[width=1.0\linewidth]{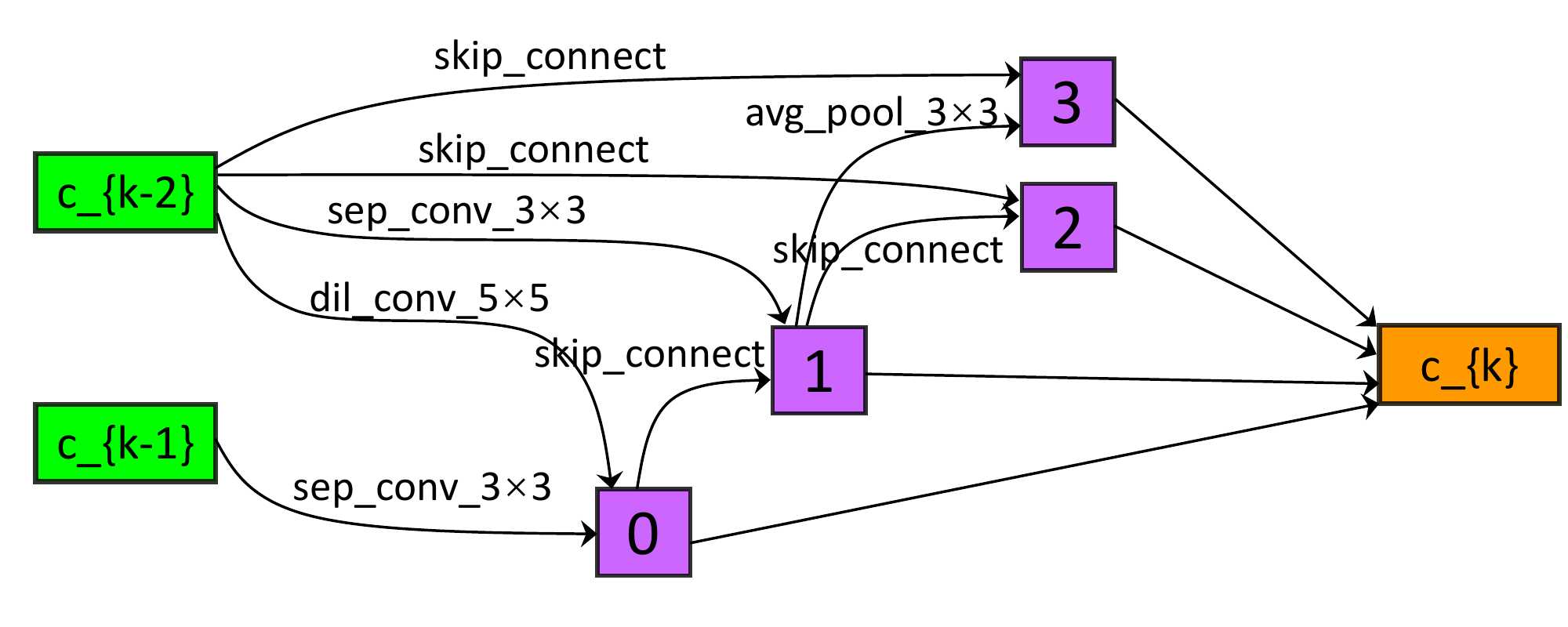}}\label{ncells_dv2}
\end{center}
\end{minipage}
\caption{The normal cells found on CIFAR10 with different balancing coefficients. The balancing coefficients $\lambda$ are $0.00$, $0.10$, $0.15$ and $0.20$, respectively. Latency optimization is added upon PC-DARTS, and $\lambda=0.00$ is the same as the original PC-DARTS. The latency here is measured on CIFAR10}
\label{fig:cells}
\end{figure}

\vspace{0.2cm}
\noindent$\bullet$\quad\textbf{The Impact of the Balancing Coefficient}

The balancing coefficient $\lambda$ is an important factor to control the impact of latency constraint, which directly determines the latency of the searched architectures. To show the impact of $\lambda$, different $\lambda$s are adopted to balance the performance and latency of the searched architectures. In this experiment, we set PC-DARTS~\cite{xu2019pc} as the baseline method ($\lambda=0.00$) and choose $\lambda=0.10$, $\lambda=0.15$ and $\lambda=0.20$ to conduct three independent search runs. The normal cells of the searched architectures and their corresponding latency and test errors are shown in Figure~\ref{fig:cells}. With the increase of $\lambda$, the latency of the searched architectures is reduced while the performance is relatively stable. It means that our latency optimization can effectively decrease the latency without affecting the searched performance. However, if we continue to increase $\lambda$ to be larger than $0.2$, parameter-free operations will dominate the searched architectures and thus much larger test errors are reported. 

\vspace{0.2cm}
\noindent$\bullet$\quad\textbf{Robustness to Latency Prediction Error}

As shown in Section~\ref{approach:latency}, The latency prediction module (LPM) still suffers an Absolute Error of $0.82$ (ms). We perform additional experiments to demonstrate that it is enough to offer a good latency constraint with an LPM of such precision and the framework is robust to the latency prediction error. A random noise with a distribution of $\mathcal{N}(0,0.025)$ is added on the predicted latency. We compare the latency of the searched architectures with LPM constraint when $\lambda=0.10$ and $\lambda=0.20$. As shown in Table.~\ref{table.3}, with the injected noise, the LPM still effectively guides to search the latency aware architectures under different balancing coefficients, which shows the robustness of the proposed LPM and latency-aware architecture framework.

\begin{table}[!t]
\begin{center}
\caption{\textbf{Left}: latency-aware architecture search with added noise. \textbf{Right}: comparing latency-aware search to FLOPs-aware search. Here, $\eta$ and $\lambda$ are the balancing coefficients for FLOPs-aware architecture search and latency-aware architecture search, respectively. All numbers (FLOPs and latency) are measured on CIFAR10 using an NVIDIA Tesla-P100 GPU}
\label{table.3}
\begin{tabular}{|l||c|c|c||l||c|c|c|}
\hline
Methods & $\lambda$ & Latency & Test Error & Methods & $\eta/\lambda$ & FLOPs & Latency  \\
\hline\hline
\multirow{2}*{LPM w/o noise} & 0.1 & 35.5ms & 2.64\% & \multirow{2}*{FLOPs-aware} & 0.005 & 533M & 29.1ms \\
\cline{2-4}\cline{6-8}
{} & 0.2 & 27.7ms & 2.61\% & {} & 0.007 & 462M & 25.2ms \\
\hline
\multirow{2}*{LPM w/ noise} & 0.1 & 36.1ms & 2.69\% & \multirow{2}*{Latency-aware} & 0.100 & 551M & 28.0ms \\
\cline{2-4}\cline{6-8}
{} & 0.2 & 28.3ms & 2.72\% & {} & 0.200 & 460M & 23.2ms \\
\hline
\end{tabular}
\end{center}
\end{table}

\vspace{0.2cm}
\noindent$\bullet$\quad\textbf{Comparison to FLOPs-Aware Architecture Search}

To show the effectiveness of latency-aware architecture search, we conduct FLOPs-aware architecture search as the control group. Different from the latency of an architecture, FLOPs is irrelevant to the route of connections but the operation itself. It is easy to apply the FLOPs constraint as a differentiable term. We measure the FLOPs of each
operation in the search space and use a lookup table to compute the overall FLOPs by adding up the FLOPs of each involved operation. A balancing coefficient $\eta$ is adopted to balance performance and FLOPs in the search scenario. We conduct two independent FLOPs-aware architecture search with $\eta=0.005$ and $\eta=0.007$ and the latency of the discovered architectures is compared with the architectures searched by latency-aware architecture search with $\lambda=0.100$ and $\lambda=0.200$. The result shows that the latency-aware architecture search approach can discover architectures with lower latency than the FLOPs-aware approach when the searched architectures have comparable FLOPs.

\begin{table}[!t]
\centering
\caption{Comparison with state-of-the-art architectures on ImageNet (mobile setting). Latency is measured on one Tesla-P100 GPU with a batch size of $32$ and an input size of $224\times224$. In latency-aware approaches, training the LPM requires additional $0.4$ GPU-days}
\label{tab.image}
\small
\begin{threeparttable}[b]
\resizebox{\textwidth}{!}{
\begin{tabular}{lccccccc}
\hline
\textbf{\multirow{2}{*}{Architecture}} & \multicolumn{2}{c}{\textbf{Test Err. (\%)}} & \textbf{Params} & $\times+$ & \textbf{Latency}& \textbf{Search Cost} & \textbf{\multirow{2}{*}{Search Method}} \\
\cmidrule(lr){2-3}
&                            \textbf{top-1} & \textbf{top-5} & \textbf{(M)} & \textbf{(M)}& \textbf{(ms)} & \textbf{(GPU-days)} &\\
\hline
Inception-v1~\cite{szegedy2015going}          & 30.2 & 10.1 & 6.6 & 1448 &- & -    & manual \\
MobileNet~\cite{howard2017mobilenets}         & 29.4 & 10.5 & 4.2 & 569  &- & -    & manual \\
MobileNet 1.4$\times$ (v2)~\cite{Sandler2018MobileNetV2IR}    & 25.3 & - & 6.9 & 585  &27.7 & -    & manual \\
ShuffleNet 2$\times$ (v1)~\cite{zhang2018shufflenet} & 26.4 & 10.2 & $\sim$5  & 524  &- & -    & manual \\
ShuffleNet 2$\times$ (v2)~\cite{ma2018shufflenet}    & 25.1 & - & $\sim$5  & 591  &- & -    & manual \\
\hline
NASNet-A~\cite{zoph2018learning}              & 26.0 & 8.4  & 5.3 & 564  & 48.7& 1800 & RL \\
NASNet-B~\cite{zoph2018learning}              & 27.2 & 8.7  & 5.3 & 488  &- & 1800 & RL \\
NASNet-C~\cite{zoph2018learning}              & 27.5 & 9.0  & 4.9 & 558  &- & 1800 & RL \\
AmoebaNet-A~\cite{real2018regularized}        & 25.5 & 8.0  & 5.1 & 555  &- & 3150 & evolution \\
AmoebaNet-B~\cite{real2018regularized}        & 26.0 & 8.5  & 5.3 & 555  &- & 3150 & evolution \\
AmoebaNet-C~\cite{real2018regularized}        & 24.3 & 7.6  & 6.4 & 570  &- & 3150 & evolution \\
PNAS~\cite{liu2018progressive}                & 25.8 & 8.1  & 5.1 & 588  &47.3& 225  & SMBO \\
MnasNet-92~\cite{tan2018mnasnet}              & 25.2 & 8.0  & 4.4 & 388  &- & -    & RL \\
\hline
SNAS (mild)~\cite{xie2018snas}     & 27.3 & 9.2  & 4.3 & 522  & 23.0& 1.5  & gradient-based \\
ProxylessNAS (GPU)$^\ddagger$~\cite{cai2018proxylessnas}       & 24.9 & 7.5  & 7.1 & 465  &- & 8.3  & gradient-based \\
BayesNAS~\cite{zhou2019bayesnas}                              & 26.5 & 8.9  & 3.9 & -    &- & 0.2  & gradient-based \\
GDAS~\cite{dong2019searching}                         & 26.0 & 8.5  & 5.3 & 581  &32.2 & 0.3 & gradient-based \\
\hline
DARTS (2nd order)~\cite{liu2018darts}      & 26.7 & 8.7  & 4.7 & 574  & 28.5& 0.3    & gradient-based \\
LA-DARTS                                                 & 25.2&  8.0 & 5.1 & 575 & 26.2 & 0.3$+$0.4 & gradient-based\\
\hline
P-DARTS (CIFAR10)~\cite{chen2019progressive}                 & 24.4 & 7.4  & 4.9 & 557  & 29.0 & 0.3  & gradient-based \\
LA-P-DARTS (CIFAR10)& 24.6 & 7.4  & 4.8 & 550  & 27.1 & 0.3$+$0.4  & gradient-based \\
\hline
PC-DARTS (CIFAR10)~\cite{xu2019pc}                          & 25.1 & 7.8  & 5.3 & 586  &31.7 & 0.1 & gradient-based \\
LA-PC-DARTS (CIFAR10)                                         & 24.9 & 7.9  & 5.3 & 598  &26.1 & 0.1$+$0.4 & gradient-based\\
\hline
\end{tabular}
}

\end{threeparttable}
\end{table}

\subsection{Experiments on ImageNet}
\label{experiments:ImageNet}

The ILSVRC2012~\cite{deng2009imagenet}, a subset of ImageNet, is used to test the transferability of architectures discovered on CIFAR10. The ILSVRC2012 consists of $1\rm{,}000$ object categories and $1.28\mathrm{M}$ training and $50\mathrm{K}$ validation images for recognition task. All images are of high-resolution and roughly equally distributed over all classes. Following the conventions~\cite{zoph2018learning,liu2018darts,xu2019pc}, we apply the \textit{mobile setting} where the input image size is fixed to be $224\times224$ and the number of multi-add operations does not exceed $600\mathrm{M}$ in the testing stage.


The evaluation on ILSVRC2012 follows DARTS, P-DARTS, and PC-DARTS, which also starts with three convolution layers of stride $2$ to reduce the resolution of feature maps from $224\times224$ of the input images to $28\times28$. $14$ cells ($12$ normal cells and $2$ reduction cells) are stacked beyond this point. The network is trained from scratch for $250$ epochs using a batch size of $1\rm{,}024$ on 8 Tesla V100 GPUs. The network parameters are optimized using an SGD optimizer with a momentum of $0.9$, an initial learning rate of $0.5$ (decayed down to zero linearly), and a weight decay of $3\times10^{-5}$. Additional enhancements are adopted including label smoothing and an auxiliary loss tower during training. Learning rate warm-up is applied for the first $5$ epochs. The latency is measured following the same setting used on CIFAR10.

As shown in Table~\ref{tab.image}, with approximately the same FLOPs, LA-DARTS has a $19\%$ lower latency than the original DARTS. Also, the latency of LA-P-DARTS and LA-PC-DARTS is $27.1\mathrm{ms}$ and $26.1\mathrm{ms}$, $7\%$ and $18\%$ lower than the original version, respectively, while the accuracy of the searched architectures is not impacted (within an acceptable range of $\pm0.2\%$). In the future, with a larger search space, we expect that our algorithm has larger room of improvement in reducing the network latency.

\subsection{Transplanting to CPU}
\label{experiments:CPU}

Last but not least, we transplant the proposed pipeline to search for efficient architectures on an Intel E5-1620 CPU. We use the LPM trained and evaluate in Section~\ref{approach:latency} which, with $80\mathrm{K}$ training architectures, reports an absolute error of $8.27\mathrm{ms}$ and a relative error of $5.32\%$ (see Table~\ref{table.1}). We use this LPM to replace the one used in Section~\ref{experiments:CIFAR10}, and adjust the balancing coefficient, $\lambda$, into smaller values since the latency on CPU is often much larger.

\begin{figure}[t]
\centering
\begin{minipage}{0.49\textwidth}
\subfloat[$\lambda=0.025$, Lat.: $122.1\mathrm{ms}$, Err.: $2.75\%$]{\includegraphics[width=1.0\linewidth]{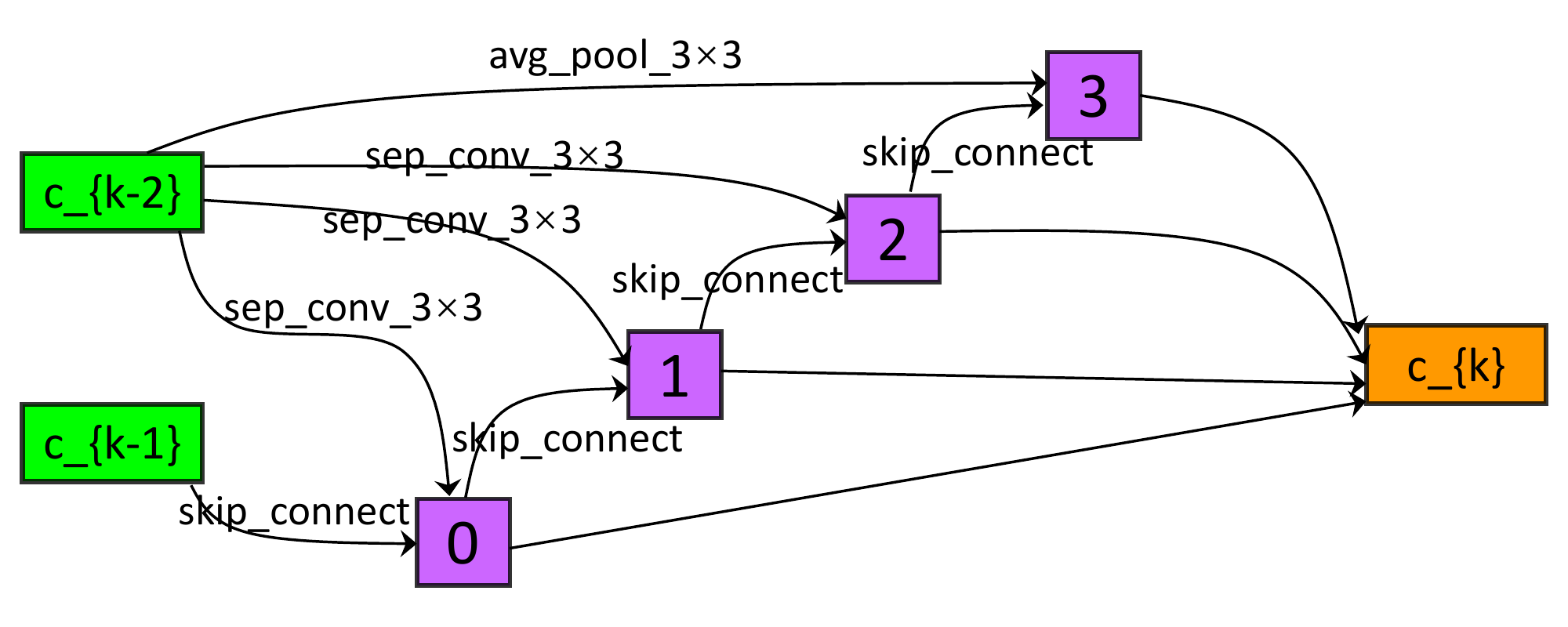}}\label{ncells_s1}\\
\end{minipage}
\begin{minipage}{0.49\textwidth}
\subfloat[$\lambda=0.015$, Lat.: $148.3\mathrm{ms}$, Err.: $2.65\%$]{\includegraphics[width=1.0\linewidth]{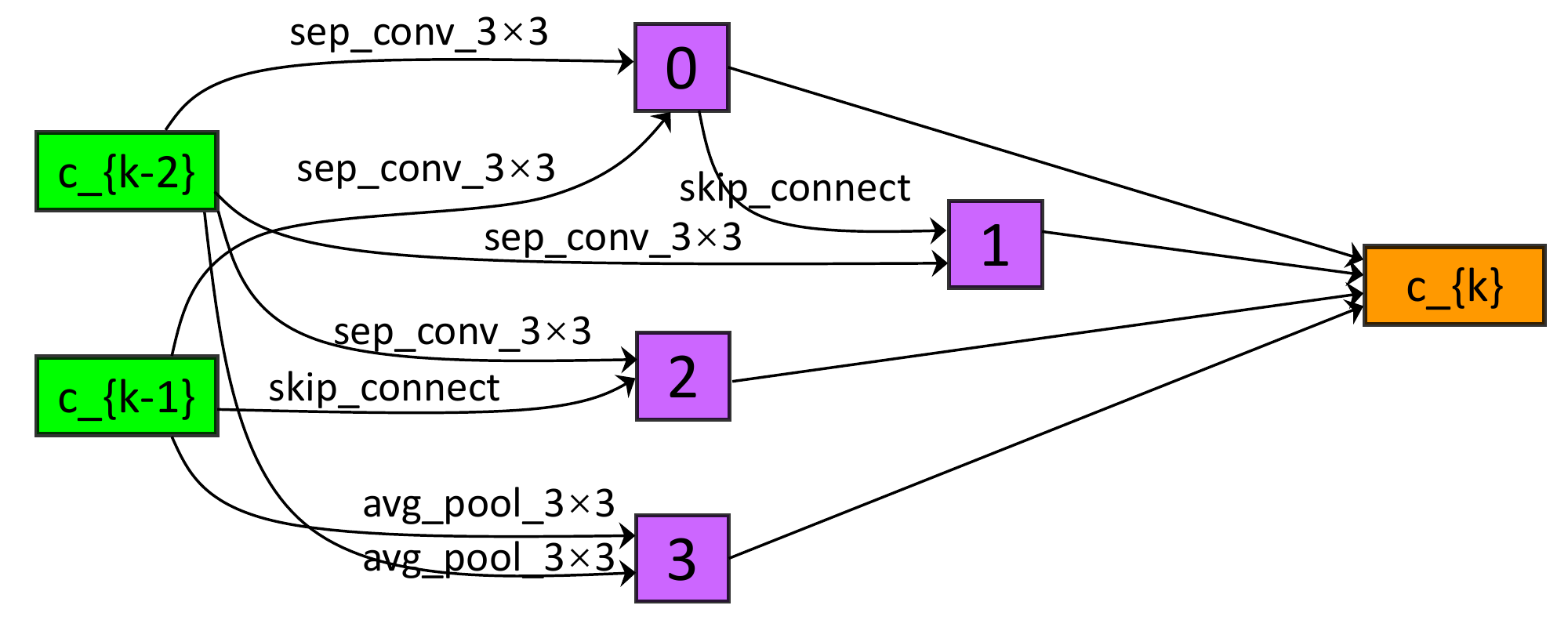}}\label{ncells_s3}\\
\end{minipage}
\caption{The normal cells found on CIFAR10 with latency-aware search on CPU. We use PC-DARTS with different balancing coefficients, and ${\lambda}={0}$ leads to the architecture shown in Figure~\ref{fig:cells} (a)}
\label{fig:cells2}
\end{figure}

We use PC-DARTS to search on CIFAR10. With two balancing coefficients, ${\lambda}={0.025}$ and ${\lambda}={0.015}$, we obtain two architectures denoted by LA-PC-DARTS-A and LA-PC-DARTS-B, respectively. As shown in Table~\ref{tab.cpu}, the increase of $\lambda$ leads to reduced latency as well as performance of the searched architecture, which is the same as searching in GPU. Compared with the original PC-DARTS, LA-PC-DARTS-B enjoys a nearly $30\%$ advantage in CPU latency while reporting comparable accuracy. LA-PC-DARTS-A runs $40\%$ faster in CPU with $0.1\%$ accuracy drop. We continue evaluating LA-PC-DARTS-B on ILSVRC2012 and obtain a $25.1\%$ top-1 test error, the same as the original PC-DARTS, yet the CPU latency ($114.1\mathrm{ms}$ on ILSVRC2012) is 30$\%$ lower than that of PC-DARTS (164.1ms). The normal cells of LA-PC-DARTS-A and LA-PC-DARTS-B are shown in Figure~\ref{fig:cells2}.

Table~\ref{tab.cpu} also implies that CPU and GPU prefer different architectures. In particular, the architecture found on GPU is faster than LA-PC-DARTS-A on GPU, but slower on CPU. To further investigate the difference, we sample $2\mathrm{K}$ architectures from the testing set of LPM, and find the Kendall-$\tau$ coefficient between the ground-truth CPU and GPU latency is only $0.37$ ($69\%$ relative rankings are consistent). We believe such inconsistency are caused by hardware factors -- fortunately, with our approach, one can obtain efficient architectures on different devices without knowing much about them: the coefficient between the prediction and ground-truth latency of GPU is $0.83$ ($92\%$ consistent), and for CPU, $0.75$ ($87\%$ consistent), both of which are accurate enough to find efficient architectures.

\begin{table}[!t]
\small
\centering
\caption{Results of latency-aware search using PC-DARTS on CIFAR10, with LPM trained on an NVIDIA Tesla-P100 GPU and an Intel E5-1620 CPU, respectively. C-Latency and G-Latency are measured on the same CPU and GPU}
\label{tab.cpu}
\begin{tabular}{lccccc}
\hline
\textbf{\multirow{2}{*}{Architecture}} & \textbf{Test Err.}  & \textbf{Params} & \textbf{C-Latency}& \textbf{G-Latency} \\
&     \textbf{ (\%)}                    & \textbf{(M)}& \textbf{(ms)}& \textbf{(ms)} \\
\hline
PC-DARTS~\cite{xu2019pc} & 2.57&  3.6 & 208.3&40.7\\
LA-PC-DARTS (GPU) & 2.61&  2.6 &  136.7&27.7\\
\hline
LA-PC-DARTS-A (CPU)  & 2.75 & 2.8  & 122.1&30.5\\
LA-PC-DARTS-B (CPU)  & 2.65 & 3.2  & 148.3&37.3\\
\hline
\end{tabular}
\end{table}

\section{Conclusions}
\label{conclusions}

This paper presented a differentiable method for predicting the latency of an architecture in a complicated search space, and incorporated this module into differentiable architecture search. This enables us to control the balance of recognition accuracy and inference speed. We design the latency prediction module as a multi-layer regression network, and train it by sampling a number of architectures from the pre-defined search space. Our pipeline is easily transplanted to a wide range of hardware/software configurations, and helps to design machine-friendly architectures.

Our work sheds light for future research on this direction. As researchers continue exploring larger spaces of NAS, it will be more and more difficult for non-differentiable search methods to converge in reasonable search time. Also, a larger search space will also provide larger room of optimizing latency, as well as other non-differentiable factors such as power consumption, of the searched architecture. We thus expect more efforts beyond this preliminary work.

\vspace{0.2cm}
\noindent
\textbf{Acknowledgements}\quad We thank Longhui Wei, Zhengsu Chen, An Xiao, Lanfei Wang, and Kaifeng Bi for instructive discussions.

\bibliographystyle{splncs04}
\bibliography{cvpr2020}
\clearpage

\begin{appendix}

\section{Visualization of Searched Architectures}

To ease the readers to reproduce our search results, here we attach all normal and reduction cells that did not appear in the main article due to the space limit.

\subsection{Reductions Cells on CIFAR-10}

\begin{figure*}[!b]
\centering

\subfloat[$\lambda=0.00$, Latency: $40.7\mathrm{ms}$, Err.: $2.57\%$]{\includegraphics[width=0.49\linewidth]{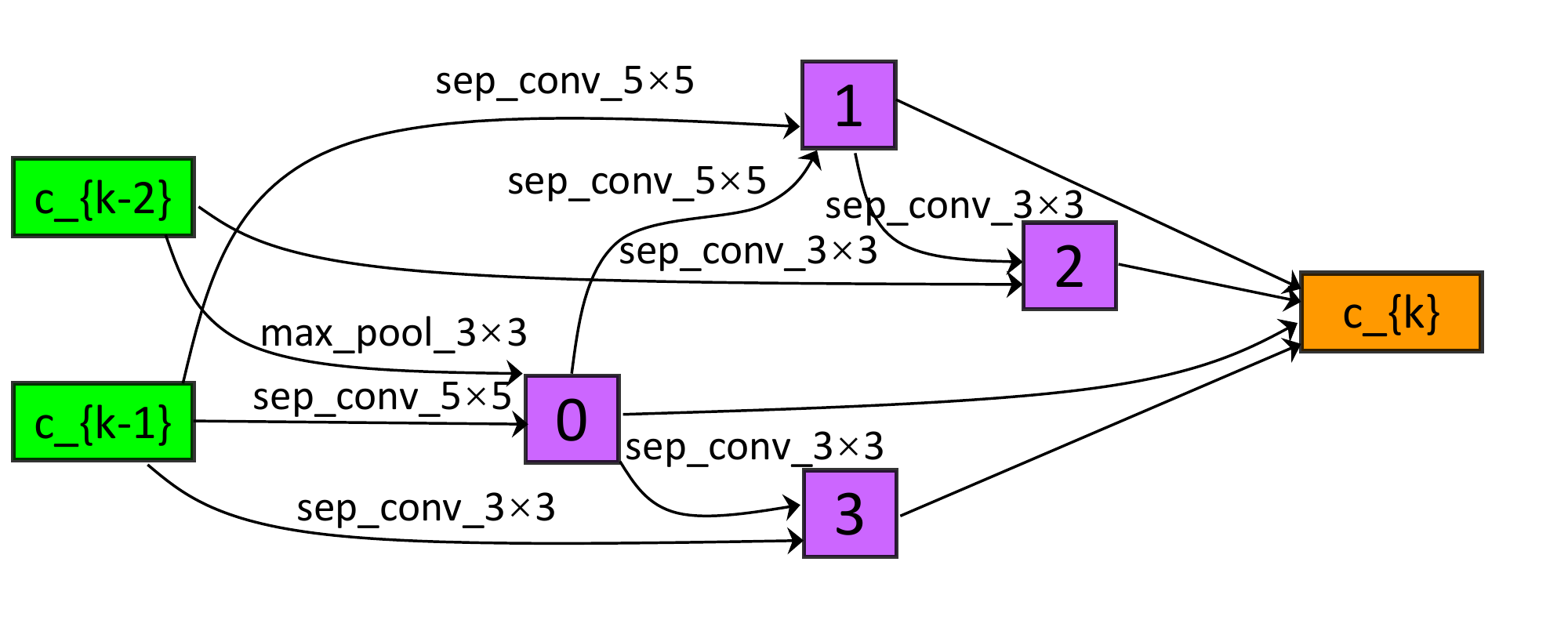}}\label{ncells_s1}
\subfloat[$\lambda=0.10$, Latency: $35.5\mathrm{ms}$, Err.: $2.64\%$]{\includegraphics[width=0.49\linewidth]{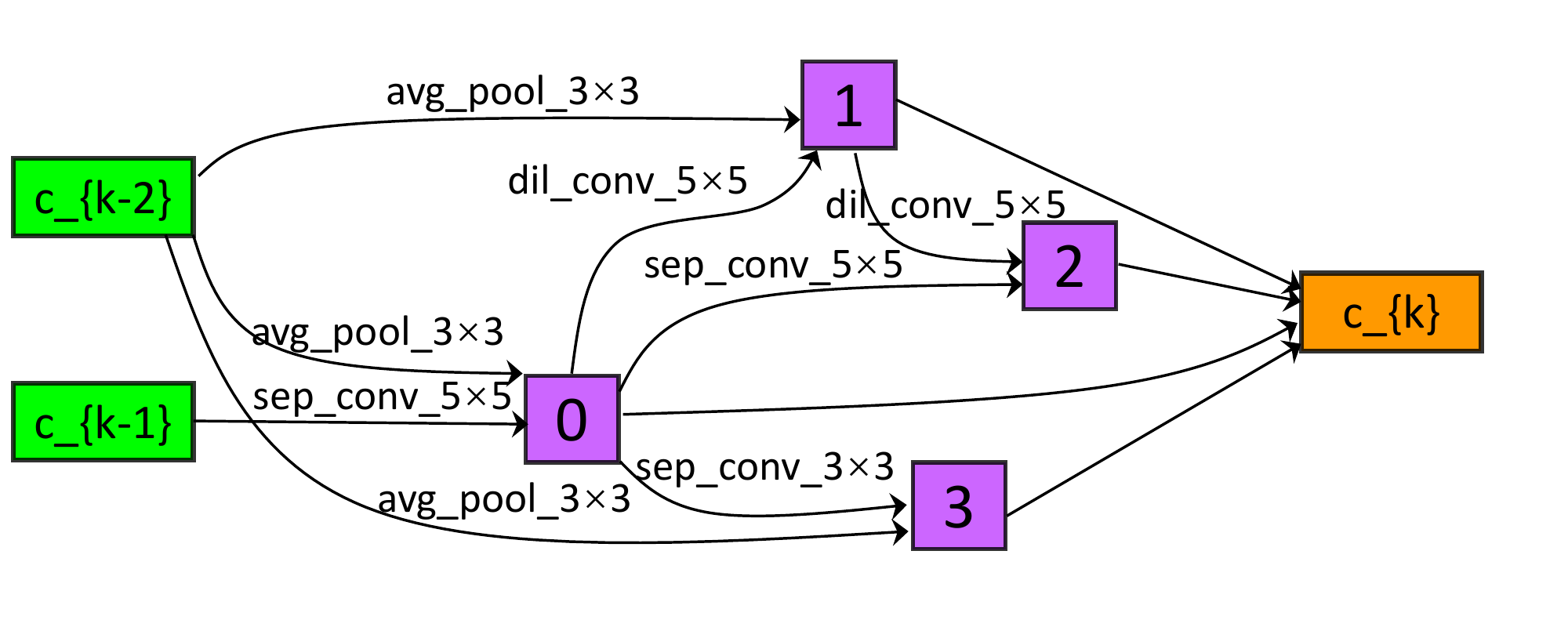}}\label{ncells_s2}\\
\subfloat[$\lambda=0.15$, Latency: $31.2\mathrm{ms}$, Err.: $2.69\%$]{\includegraphics[width=0.49\linewidth]{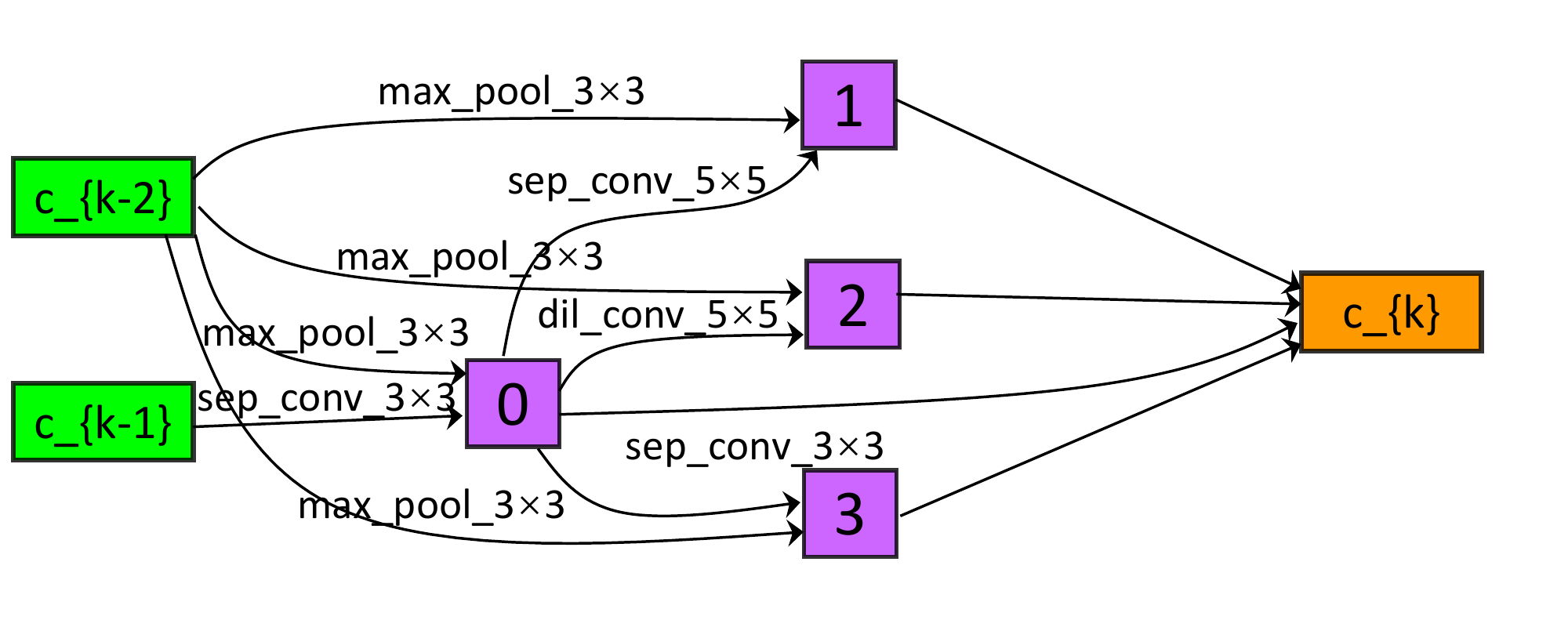}}\label{ncells_s3}
\subfloat[$\lambda=0.20$, Latency: $27.7\mathrm{ms}$, Err.: $2.61\%$]{\includegraphics[width=0.49\linewidth]{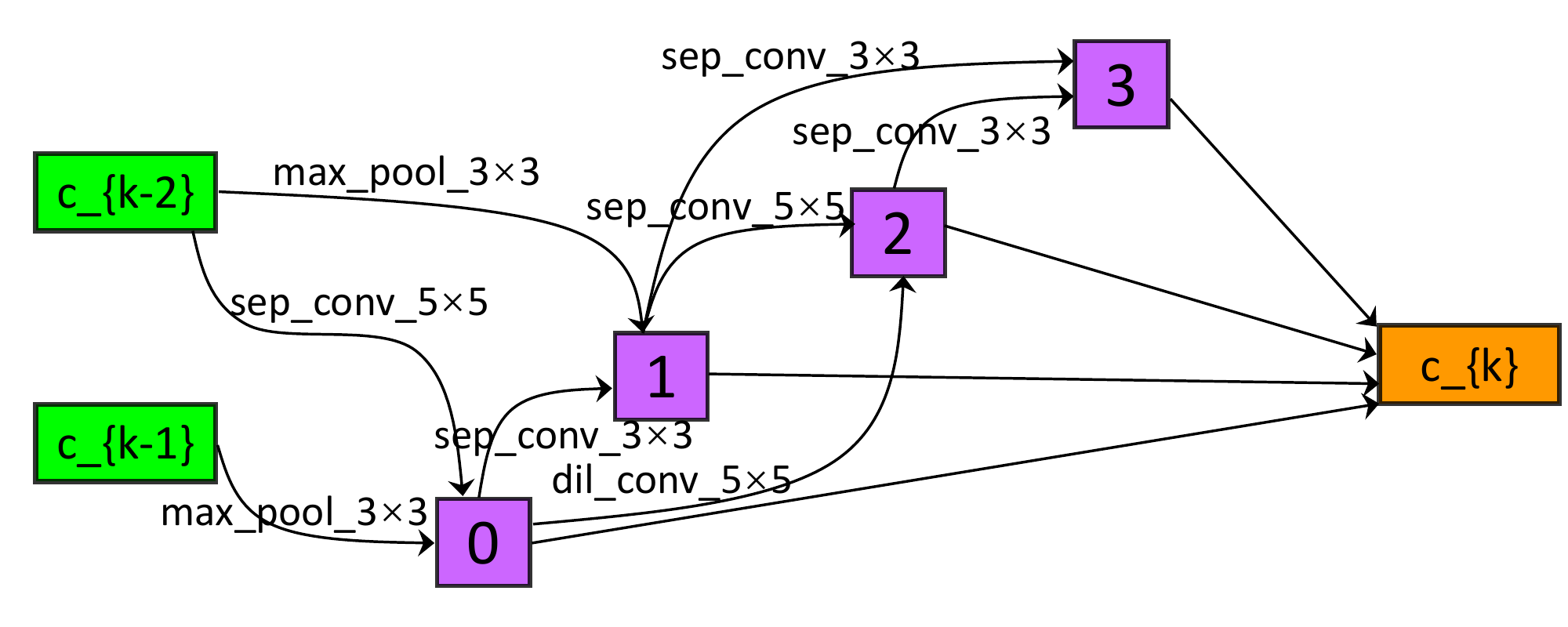}}\label{ncells_dv2}\\
\caption{The corresponding reduction cells found on CIFAR-10 with different balancing coefficients. The balancing coefficients $\lambda$ are $0.00$, $0.10$, $0.15$ and $0.20$, respectively. Latency optimization is combined with PC-DARTS and $\lambda=0.00$ is the same as the original PC-DARTS. The latency here is measured on CIFAR-10}
\label{fig:cells1}
\end{figure*}

The reduction cells of architectures found on CIFAR-10 with different balancing coefficients are shown in Figure~\ref{fig:cells1}. The balancing coefficients $\lambda$ are $0.00$, $0.10$, $0.15$ and $0.20$, respectively. Latency optimization is combined with PC-DARTS and $\lambda=0.00$ is the same as the original PC-DARTS. The latency is measured on CIFAR-10.

\subsection{Cells of LA-DARTS, LA-PC-DARTS and LA-P-DARTS}

\begin{figure*}[h]
\centering
\subfloat[The normal cell of DARTS (2nd)]{\includegraphics[width=0.49\linewidth]{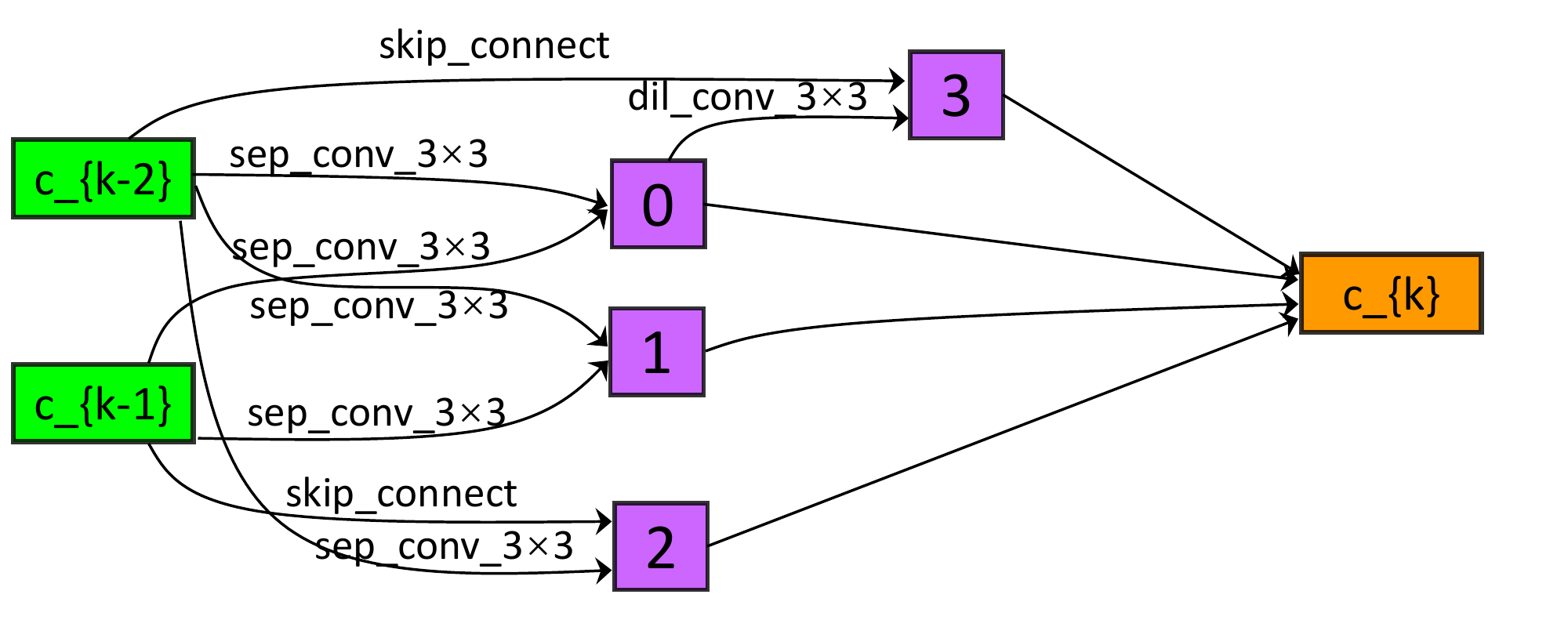}}\label{ncells_s3}
\subfloat[The reduction cell of DARTS (2nd)]{\includegraphics[width=0.49\linewidth]{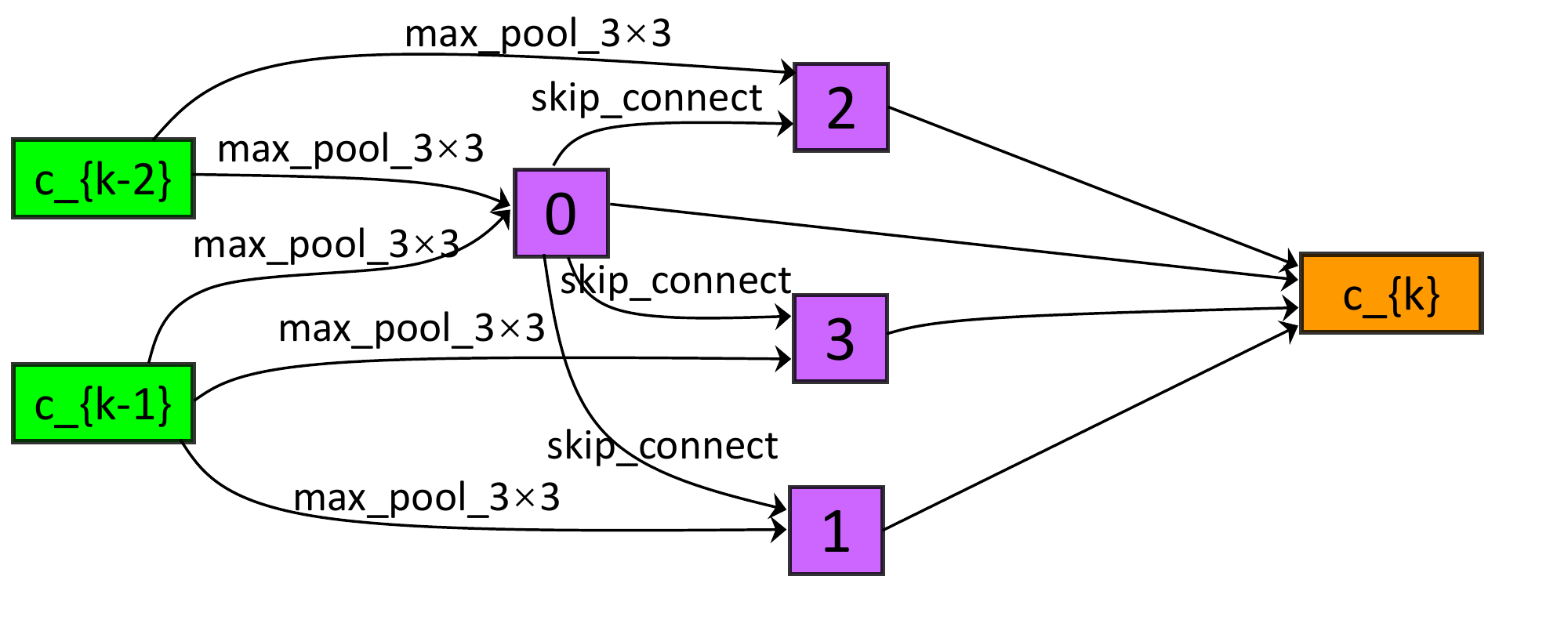}}\label{ncells_dv2}\\
\subfloat[The normal cell of PC-DARTS]{\includegraphics[width=0.49\linewidth]{cifar_1.pdf}}\label{ncells_s1}
\subfloat[The reduction cell of PC-DARTS]{\includegraphics[width=0.49\linewidth]{cifar_1r.pdf}}\label{ncells_s2}\\
\caption{Normal cells and reduction cells of DARTS (2nd) (Test error: $2.76\%$) and PC-DARTS (Test error: $2.57\%$)}
\label{fig:cello}
\end{figure*}

The normal and reduction cells of LA-DARTS, LA-PC-DARTS and LA-P-DARTS are shown in Figure~\ref{fig:cells}. The balancing coefficient $\lambda$ is $0.20$ for both LA-DARTS and LA-PC-DARTS and $0.10$ for LA-P-DARTS. Besides, the normal and reduction cells of DARTS (2nd-order) and PC-DARTS are shown in Figure \ref{fig:cello}.

\begin{figure*}[!t]
\centering
\subfloat[The normal cell of LA-DARTS]{\includegraphics[width=0.49\linewidth]{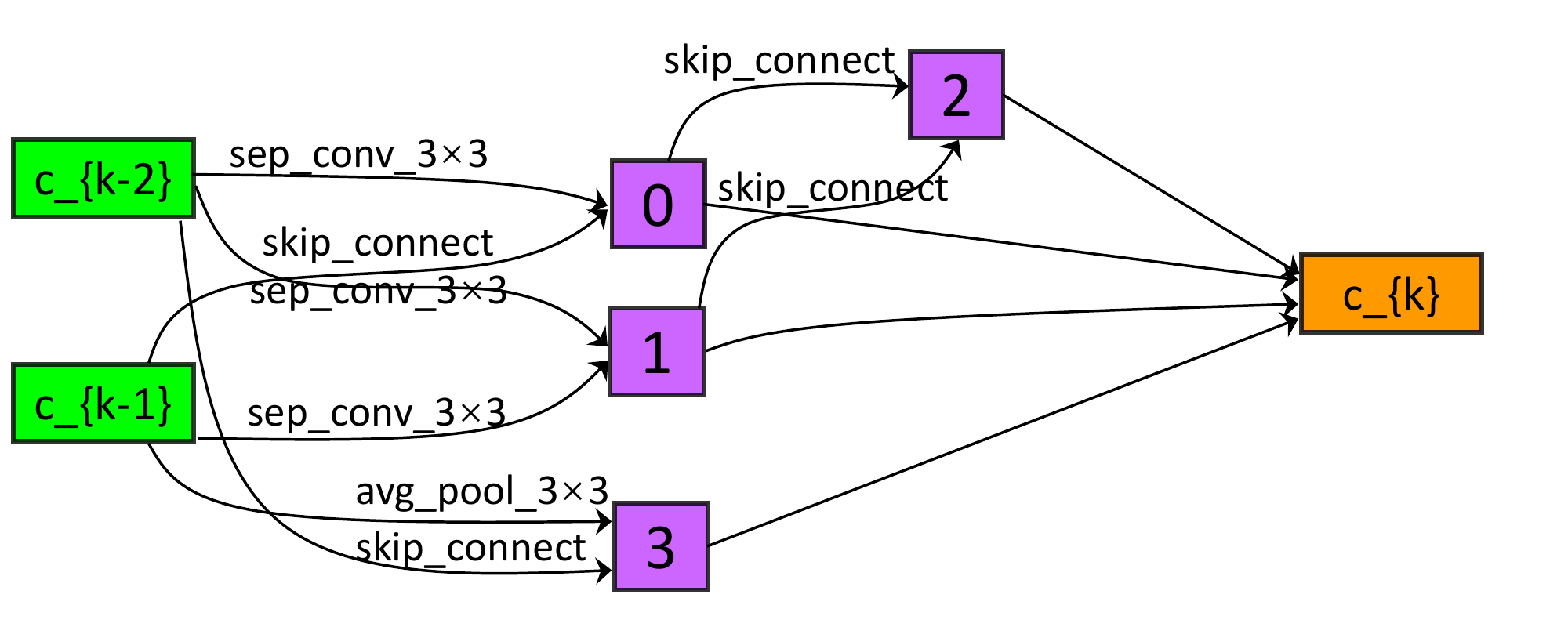}}\label{ncells_s3}
\subfloat[The reduction cell of LA-DARTS]{\includegraphics[width=0.49\linewidth]{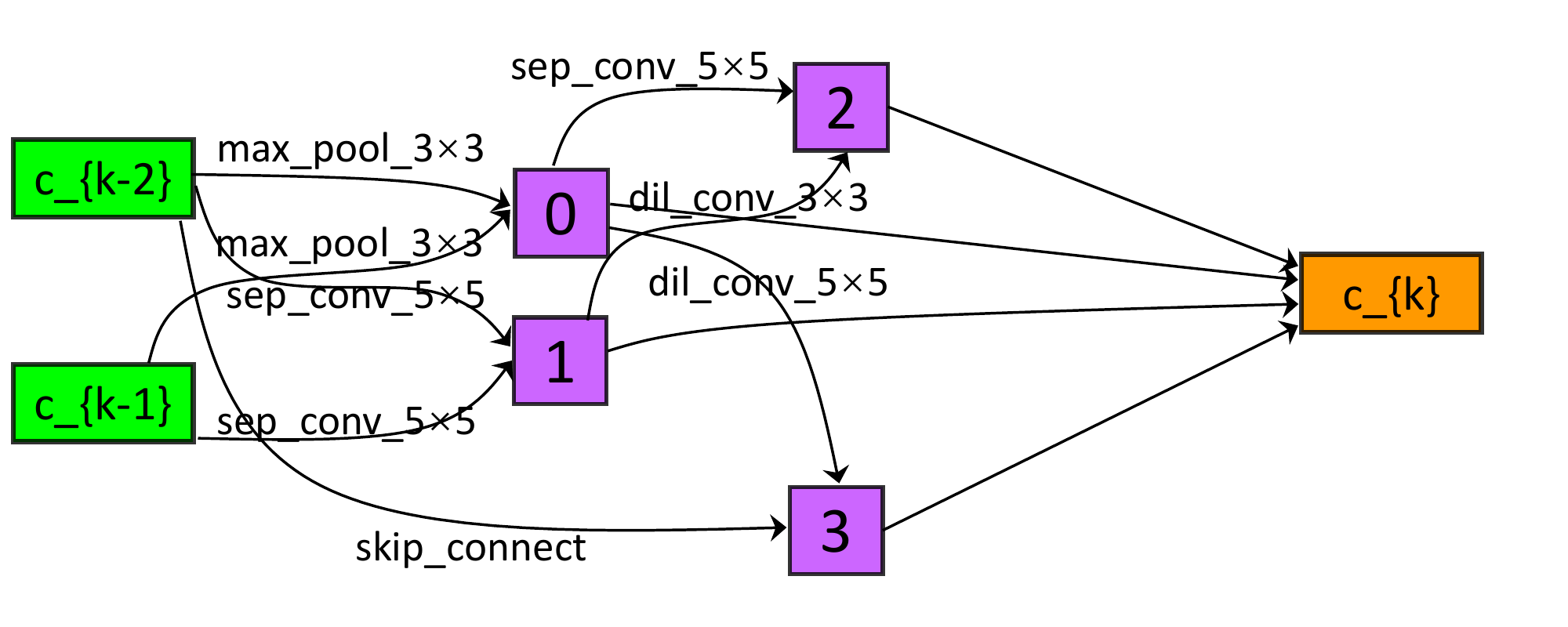}}\label{ncells_dv2}\\
\subfloat[The normal cell of LA-PC-DARTS]{\includegraphics[width=0.49\linewidth]{cifar_4.pdf}}\label{ncells_s1}
\subfloat[The reduction cell of LA-PC-DARTS]{\includegraphics[width=0.49\linewidth]{cifar_4r.pdf}}\label{ncells_s2}\\
\subfloat[The normal cell of LA-P-DARTS]{\includegraphics[width=0.49\linewidth]{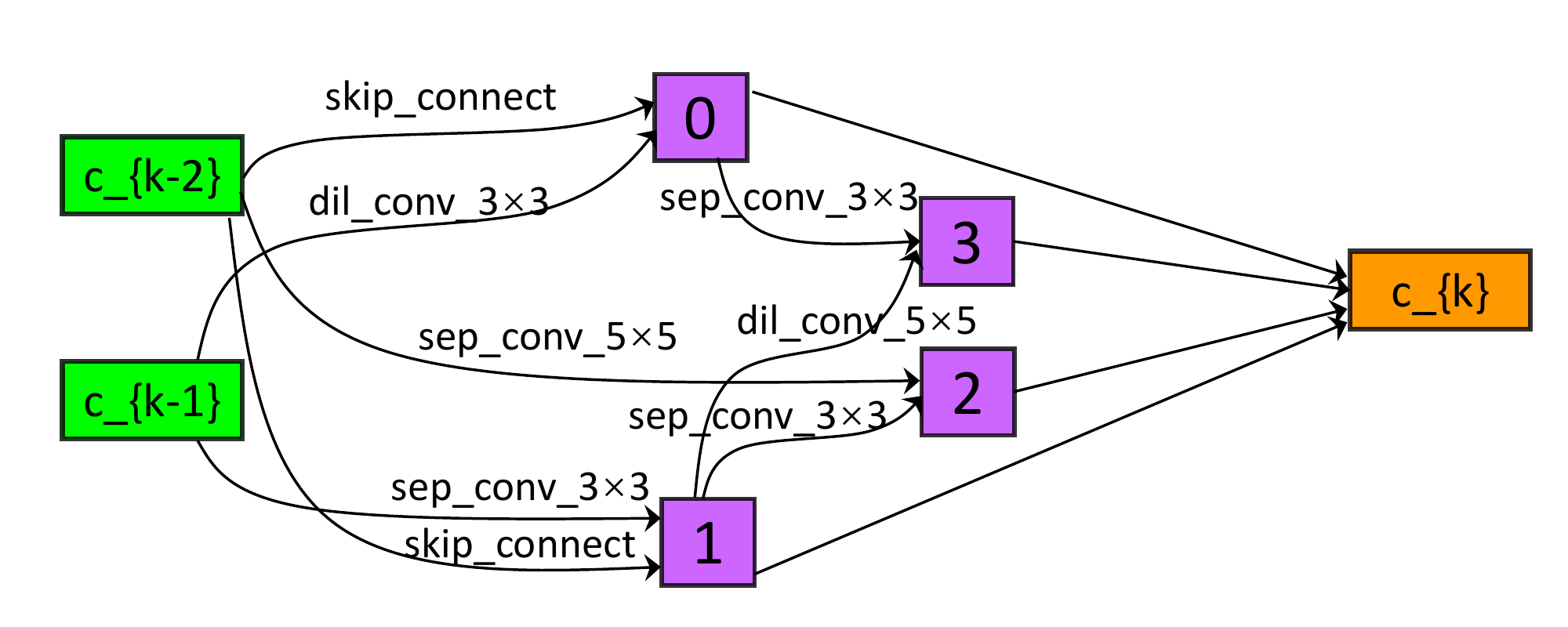}}\label{ncells_s5}
\subfloat[The reduction cell of LA-P-DARTS]{\includegraphics[width=0.49\linewidth]{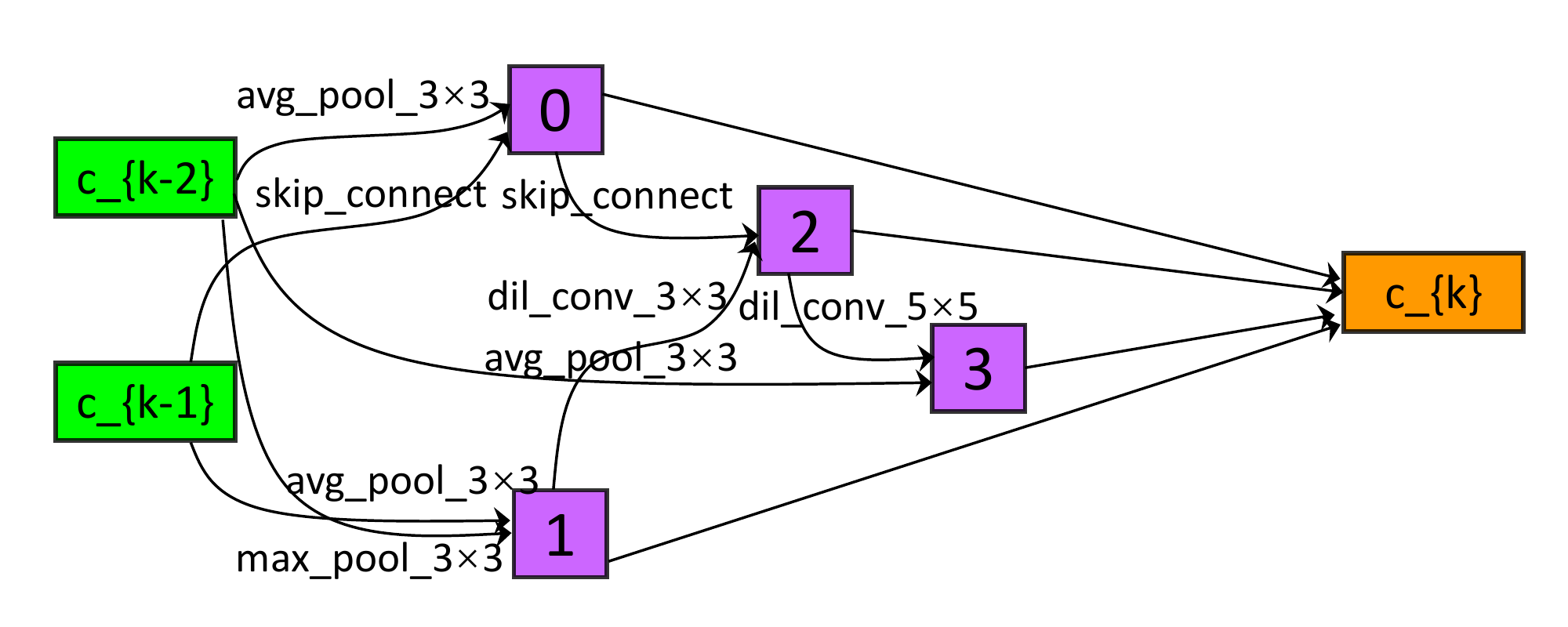}}\label{ncells_s6}\\
\caption{Normal cells and reduction cells of LA-DARTS (Test error: $2.72\%$), LA-PC-DARTS (Test error: $2.61\%$) and LA-P-DARTS (Test error: $2.52\%$)}
\label{fig:cells}
\end{figure*}

\subsection{Cells of LA-PC-DARTS-A and LA-PC-DARTS-B}

\begin{figure*}[h]
\centering
\subfloat[The normal cell of LA-PC-DARTS-A]{\includegraphics[width=0.49\linewidth]{LAPCCPU-A_n.pdf}}\label{ncells_s3}
\subfloat[The reduction cell of LA-PC-DARTS-A]{\includegraphics[width=0.49\linewidth]{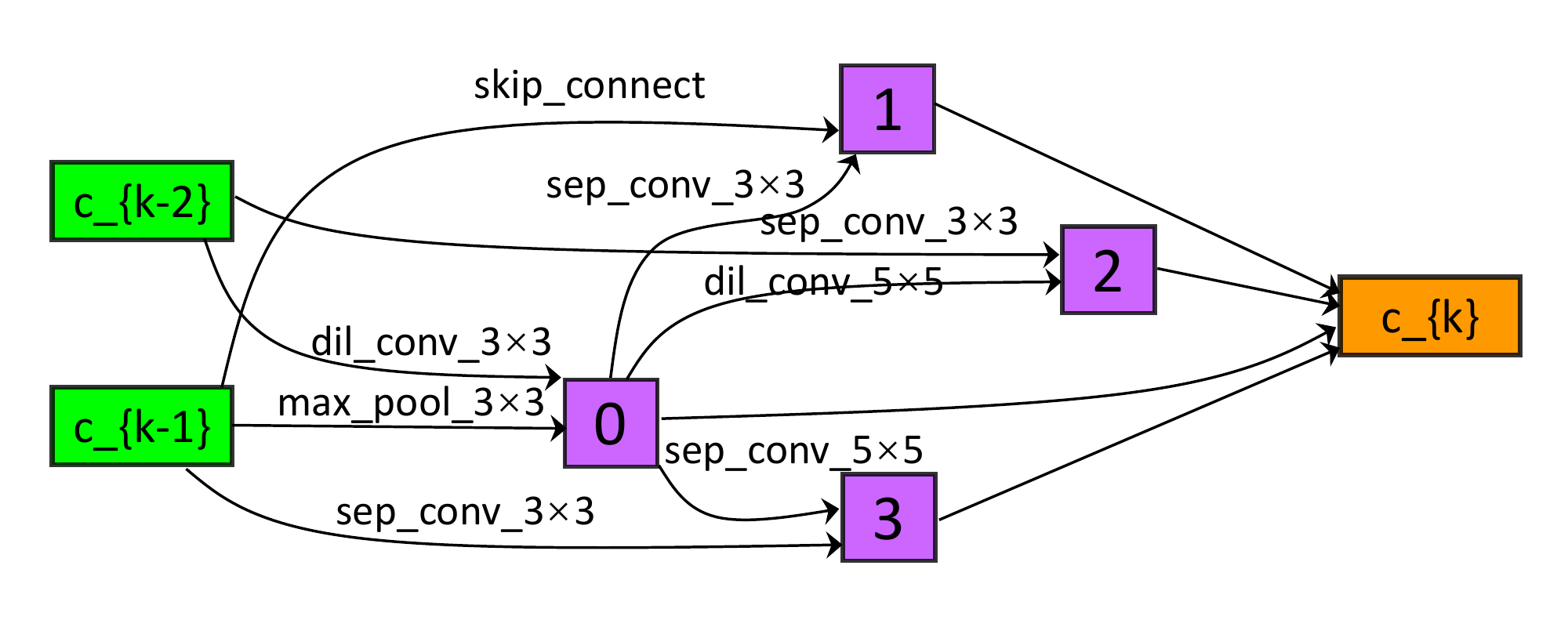}}\label{ncells_dv2}\\
\subfloat[The normal cell of LA-PC-DARTS-B]{\includegraphics[width=0.49\linewidth]{LAPCCPU-B_n.pdf}}\label{ncells_s1}
\subfloat[The reduction cell of LA-PC-DARTS-B]{\includegraphics[width=0.49\linewidth]{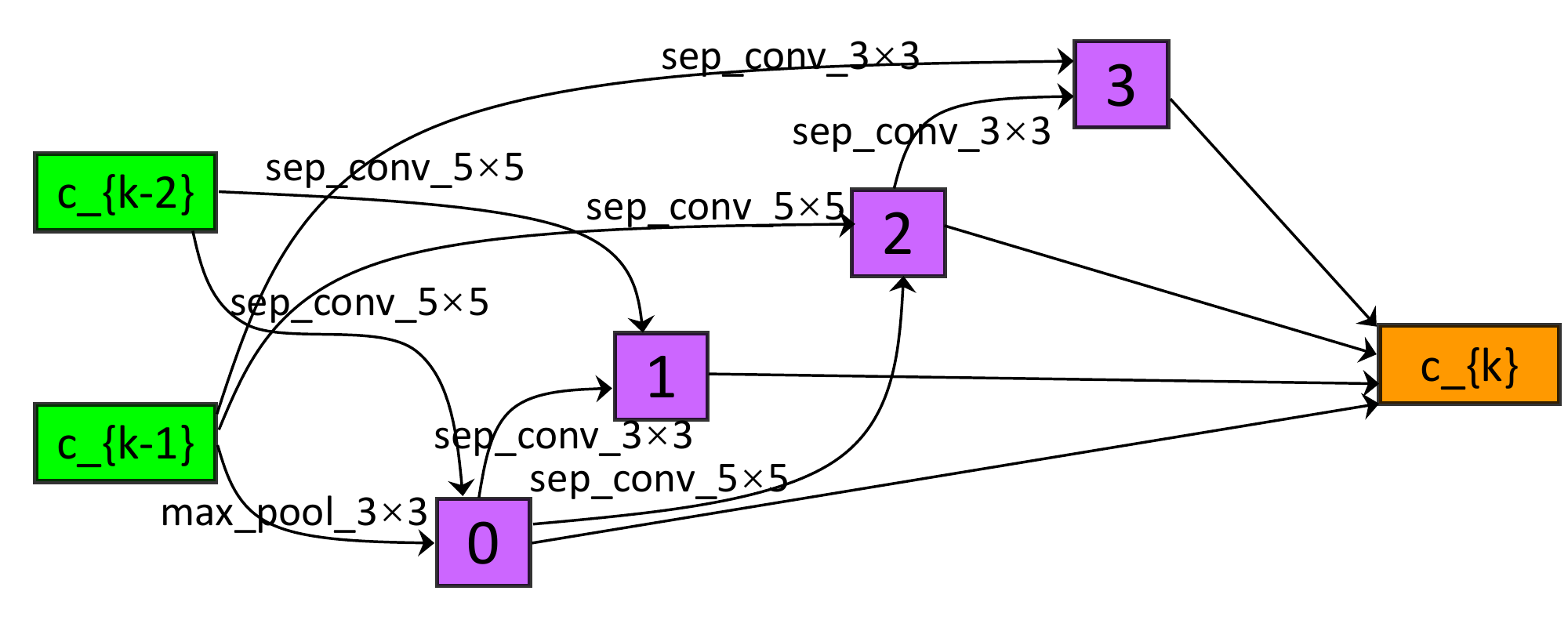}}\label{ncells_s2}\\
\caption{Normal cells and reduction cells of LA-PC-DARTS-A (Test error: $2.75\%$, CPU latency: $122.1$ms) and LA-PC-DARTS-B (Test error: $2.65\%$, CPU latency: $148.3$ms)}
\label{fig:cellcpu}
\end{figure*}

LA-PC-DARTS-A and LA-PC-DARTS-B are CPU-aware searched architectures. The normal and reduction cells of LA-PC-DARTS-A and LA-PC-DARTS-B are shown in Figure~\ref{fig:cellcpu}. The balancing coefficient $\lambda$ is $0.025$ for LA-PC-DARTS-A and $0.015$ LA-PC-DARTS-B.

\end{appendix}
\end{document}